\documentclass{article}

\usepackage{microtype}
\usepackage{graphicx}
\usepackage{subfigure}
\usepackage{booktabs}
\usepackage{hyperref}

\usepackage{algorithm}
\usepackage{algorithmic}

\usepackage{envoptarxiv}
\usepackage[round]{natbib}

\usepackage[strings]{underscore}
\usepackage{hyperref}
\usepackage{url}
\usepackage{graphicx}
\usepackage{booktabs}
\usepackage{amsmath,amssymb}
\usepackage{amsfonts}       
\usepackage{nicefrac}       
\usepackage{microtype}      
\usepackage{xcolor}         

\usepackage{enumitem}
\usepackage{graphicx}
\usepackage{sidecap}
\usepackage{diagbox}
\usepackage[small]{caption}
\usepackage{subcaption}
\usepackage{comment}
\usepackage{array}
\usepackage[export]{adjustbox}
\usepackage{float}
\usepackage{wrapfig}
\usepackage{mathtools}
\usepackage{xspace}
\usepackage{bm}
\usepackage{multirow}
\usepackage{multicol}
\usepackage[capitalize]{cleveref}
\usepackage{fvextra}
\usepackage[most]{tcolorbox}

\newcounter{tasknum}
\crefname{tasknum}{example task}{Example Tasks}
\Crefname{tasknum}{Example task}{Example tasks}

\definecolor{darkblue}{rgb}{0, 0, 0.5}

\newcommand{\framework}{\textsc{EnvOpt}\xspace}

\newif\ifhidecomments

\title{\framework: Automated Optimization for Agent Environments}
\author[1,2,3]{Mingxuan Li}
\author[1]{Huanzhi Mao}
\author[1]{Qiuyang Mang}
\author[1]{Hanchen Li}
\author[1]{Joseph E. Gonzalez}
\author[1,3]{Alex Dimakis}
\affiliation[1]{University of California, Berkeley}
\affiliation[2]{University of Chicago}
\affiliation[3]{Bespoke Labs}
\correspondence{\email{mingxuanl@uchicago.edu}}
\metadata[Code]{\url{https://github.com/itea1001/EnvOpt}}

\begin{document}

\maketitle

\begin{abstract}

An LLM agent's cost and success depend on its model, its harness, and its environment. In addition to the widely studied directions of parameter tuning and harness optimization, we propose the third direction: optimization over stationary environments. Agents are often deployed in rich environments shared by many tasks, such as a company's Slack workspace or a complex collection of data sources, documents or tickets.  We call these environments \textbf{stationary environments}. We introduce \framework, a method for automated end-to-end optimization of stationary environments for any given model and harness, using real or synthetic tasks without needing any graders or reference answers. On AutomationBench, \framework reduces the cost of Claude Haiku 4.5 by 72.7\% and raises its pass rate by 14.34\%. The same environment transfers to Opus 5, cutting its cost by 68.1\% with a 1.34\% gain. Added on top of Meta-Harness's optimized harness for Haiku 4.5, it lowers cost by a further 83.4\% and adds 11.17\% accuracy over Meta-Harness alone. Across 5 benchmarks, \framework significantly reduces the cost of agents while improving or maintaining pass rate, and is transferable to different models and different harnesses. We also discuss common successful and rejected environment optimization strategies during automated optimization. More broadly, our results show that how information is represented to an agent shapes both its cost and its success, making the environment worth optimizing in addition to the parameters and the harness.

\end{abstract}

\section{Introduction}
\label{sec:intro}

An LLM agent's cost and success rate depend on three things: the model parameters, the harness, and the structure of the environment it interacts with. The first two are actively optimized through parameter tuning and harness optimization \citep{lee2026metaharnessendtoendoptimizationmodel,agrawal2026gepareflectivepromptevolution}, while the environment is often considered immutable. Yet, the same information can be given to an agent in very different forms. For example, a company's Slack history can either sit behind a remote server and be accessible via MCP or exported in plain text files that can be read and searched with terminal tools like grep. This choice alone can significantly affect how much work the agent needs to perform to solve a task \citep{mak2026bashneedempiricalstudy}. In this paper, we propose optimizing \textbf{stationary environments} as a third axis for improving LLM agents, alongside model parameter tuning and harness optimization. Stationary environments refer to environments with large collections of artifacts that remain the same or are similar across many tasks. For example, a banking agent uses the same enterprise platforms and databases for a massive number of queries; many tasks for a company's Slack assistant depend on the same Slack workspace; feature requests for a toolkit can rely on the same code repository. We aim to optimize how the stationary environments are organized and represented to agents.

\begin{figure*}[t]
    \centering
    \includegraphics[width=1\textwidth]{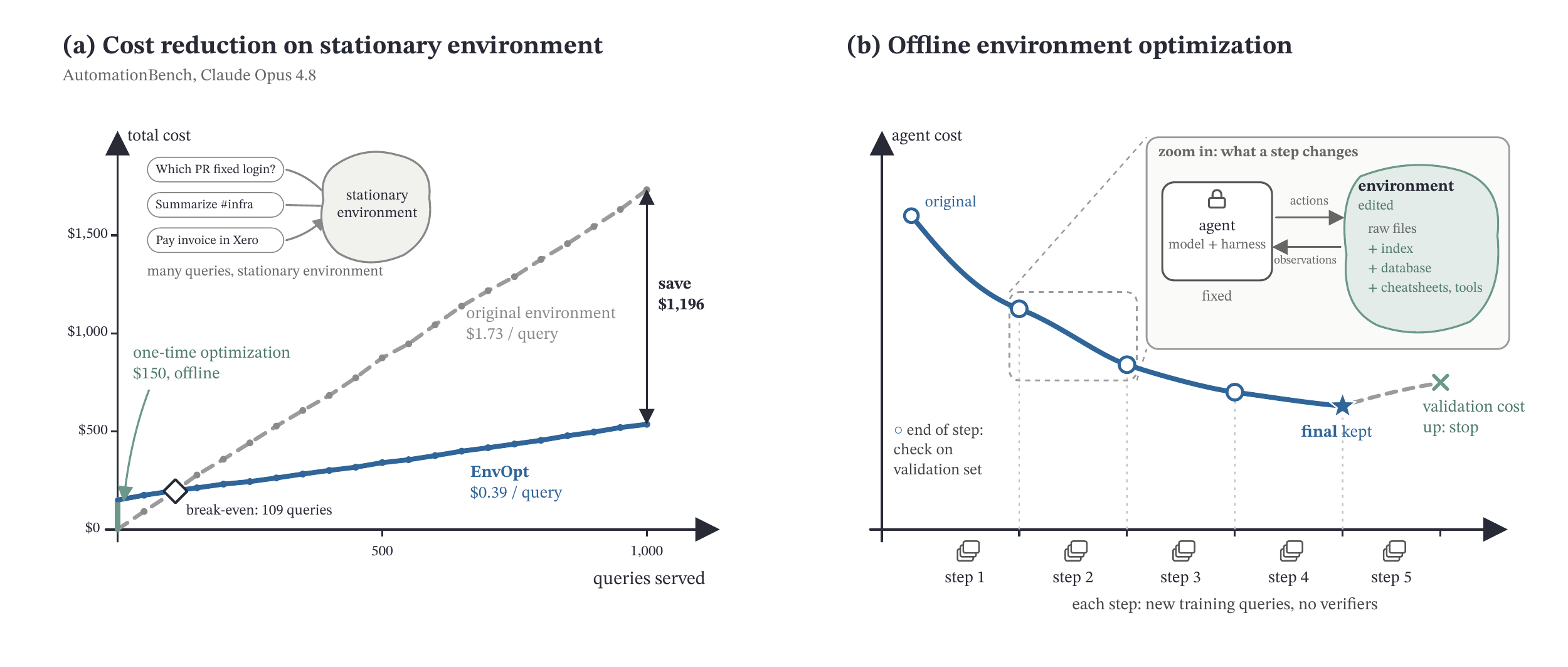}
    \caption{
    \framework on AutomationBench. \textbf{Left:} \framework costs around USD \$150 for an offline optimization of the stationary environment in AutomationBench, and saves cost of running Claude Opus 4.8 by 77.5\%, easily saving many times the cost of running \framework. Lines are derived by random sampling actual cost of running the test set tasks. \textbf{Right:} Illustration of how offline optimization works in \framework. With a fixed model and harness, \framework uses disjoint sets of training tasks (without verifiers) at each step to optimize the environment. After each step, the optimized environment is tested on a separate validation set to determine when the whole process should be stopped.
    }
    \label{fig:fig1}
\end{figure*}

\begin{figure*}[t]
    \centering
    \includegraphics[width=1\textwidth]{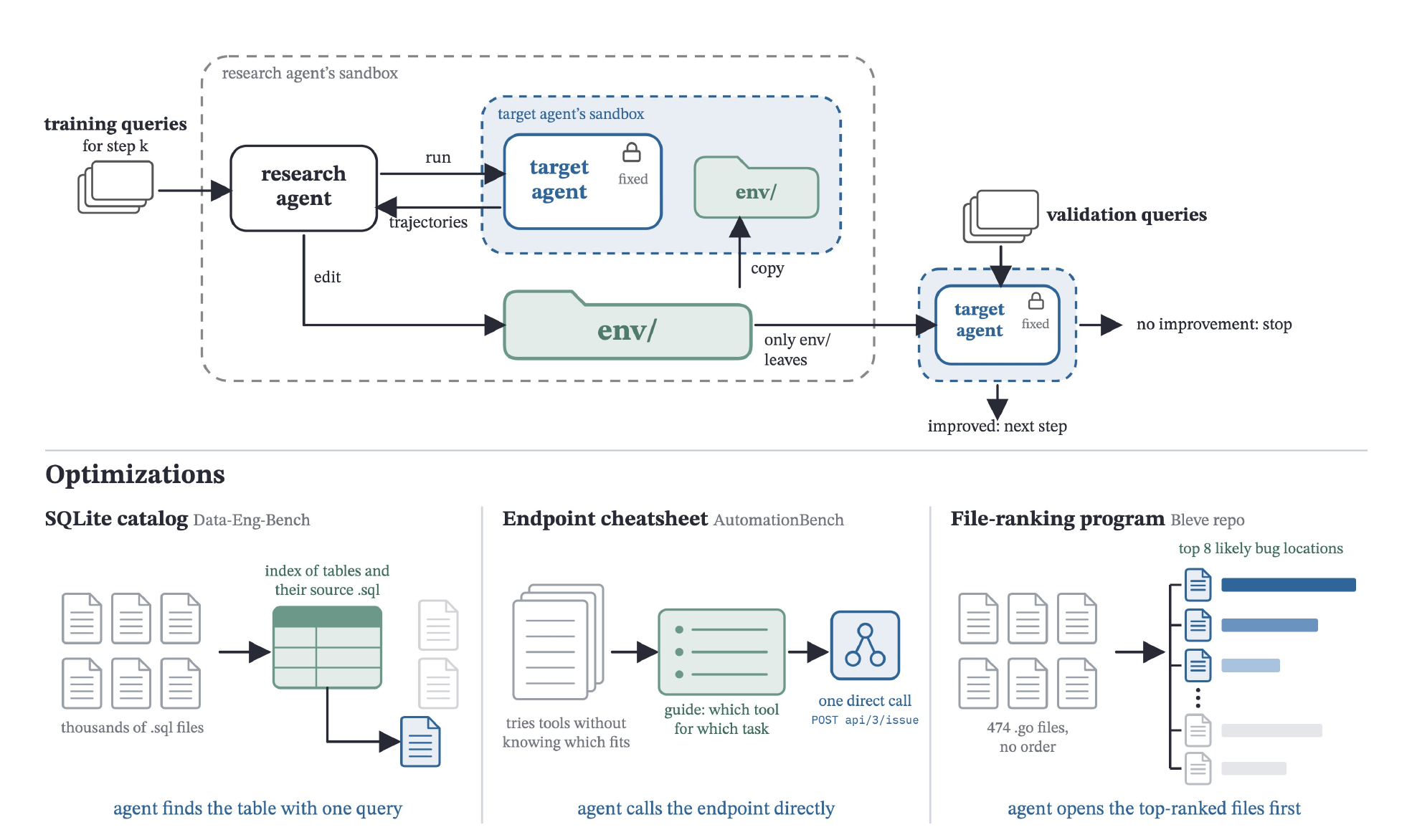}
    \caption{
    Overview of \framework methodology. \textbf{Upper:} \framework optimizes the environment in multiple steps, with each step using a disjoint set of training tasks without verifiers. In each step, a research agent is launched in a sandbox and given the original environment as a local directory and the permission to run target agent in another sandbox. The research agent optimizes environment by iteratively editing the env directory and testing the target agent. At the end of each step, we run validation queries (with verifiers) to decide if \framework has converged or should continue running. \textbf{Lower:} Examples of environment optimization across three benchmarks. The research agent constructed SQLite catalog to map tables to source files, made cheatsheet for more efficient tool use, and built ranking programs to prioritize important files for bug fixes.
    }
    \label{fig:methods}
\end{figure*}

Though the environment has not received the same systematic optimization as model parameters and harnesses, previous work has shown that refining agent-environment interactions can improve agent performance. SWE-agent \citep{yang2024sweagent} and subsequent work \citep{xu2026devilinterfaceevaluatingtool} showed that the same model fixes more issues with fewer steps when given a better interface to the repository \citep{xu2026devilinterfaceevaluatingtool}, but these interfaces were manually designed. Later work reshapes environments to create more training signal for reinforcement learning (RL) \citep{lu2026dontjustfinetuneagent,huang2026envharnessawakeningstaticworlds}, but these modifications are designed to improve model training rather than the execution efficiency of deployed agents. \textsc{PREPING} and \textsc{corpus2skill} mainly focus on distilling information from the environment into reusable agent memory and skills, rather than directly optimizing how the stationary environment itself is organized and represented \citep{choi2026prepingbuildingagentmemory,sun2026corpus2skilldistillingenterpriseknowledge}. In sum, the community has explored how to utilize environment during RL, optimize the tool interface of environment, or refine the environment for specific use cases, but environment optimization itself as the third direction of agent optimization remains underexplored.

We present \framework, which is an automated pipeline to optimize the environment for any agent and any stationary environment. Using training tasks without verifiers, we let a coding agent, equipped with access and budget to run a target agent on the stationary environment, to optimize the environment representation for the target agent. The optimization target includes what the environment holds (e.g. adding a database to the environment or restructuring it) and ways it is exposed to the target agent (e.g. new API endpoints). The optimized environment is monitored with a validation set with verifiers to track cost and pass rate, and tested on a held-out test set.

Across five benchmarks covering workspace tasks, data engineering tasks, and software engineering tasks (see \cref{sec:experiments}), \framework reduces the cost of the target agent on unseen test sets by an average of 48\%, with higher or similar pass rates. On AutomationBench \citep{shepard2026automationbench}, \framework outperforms Meta-Harness \citep{lee2026metaharnessendtoendoptimizationmodel} by 9.67\% pass rate with only 15\% of its cost. We also show that \framework constitutes an optimization direction orthogonal to harness optimization and parameter tuning in two ways: (1) the environments optimized for the target agents can transfer to other agents, even to those with different harnesses and to models stronger than the one proposing the optimizations; (2) we applied the environment from \framework on top of Meta-Harness and observe similar cost reductions and 16\% improvement in pass rate, outperforming either Meta-Harness or \framework alone.

Our contributions are as follows:

\begin{itemize}
    \item We define environment optimization over stationary environments as a new orthogonal direction for improving LLM agents. Our method can be combined with parameter tuning and harness optimization.

    \item We develop \framework, an automated pipeline for environment optimization, and show that across 5 benchmarks over 3 types of tasks, it can significantly reduce the cost of running agents on unseen tasks while improving or maintaining the pass rates.

    \item We demonstrate that \framework is indeed orthogonal to parameter tuning and harness optimization by showing strong transferability of environment optimizations to different frontier models and harnesses, and that \framework can further improve results combined with Meta-Harness.

    \item We systematically discuss the accepted and rejected design choices during running \framework, which can be used for future work on environment optimization.
\end{itemize}

\section{Method}
\label{sec:method}

\framework is an automated research pipeline that uses a coding agent such as Codex as a research agent $R$ to automatically optimize the environment representation for another target agent $T$ on tasks sharing a stationary environment.

We denote a stationary environment as $E(I)$, where $I$ is the information underlying the environment, such as the conversations in a Slack channel; $E$ is the representation of that information in a specific format, such as an official Slack workspace accessible via MCP tools or a Markdown file storing all the conversations. Our premise is that for a fixed $I$ needed for the tasks, a better representation $E^*(I)$ can significantly improve the token efficiency of agents on the tasks and, in some cases, improve accuracy as well. Given task instances $X_{\text{train}},X_{\text{val}},X_{\text{test}}\sim \mathcal{X}$ on the initial stationary environment $E_0(I)$, the objective of \framework is to find $E^*$ that minimizes the cost of the target agent $T$ on $\mathcal{X}$:

\begin{equation}
    E^{*} \;=\; \arg\min_{E} \;
    \mathbb{E}_{x \sim \mathcal{X}}
    \bigl[\mathrm{cost}\bigl(T, E(I), x\bigr)\bigr].
\end{equation}

\paragraph{Environment Evolution} As illustrated in \cref{alg:env_evolution}, \framework optimizes $E(I)$ over $N$ steps. At step $k$, the research agent $R$ is spawn in a sandbox and given a set of training tasks $X_k\subset X_{\text{train}}$, the stationary environment $E_{k-1}(I)$ from step $k-1$, the access to running the target agent $T$, and a set budget. We only provide a subset of $X_{\text{train}}$, since too much context for LLM-based optimization has been shown to be harmful \citep{li2026combeescalingpromptlearning}. We provide disjoint sets for tasks for each step to prevent overfitting. The research agent is instructed to run $T$ on $E_{k-1}(I)$, analyze the trajectories, and edit $E_{k-1}(I)$ as a directory in its filesystem to reduce the cost of $T$ on $X_k$ without hurting the pass rate (which the research agent judges on itself by proxies or answer consistency, since no verifier is provided for the training tasks as elaborated below). For example, if $E_{k-1}(I)$ is a big directory of files with no structure, $E_k(I)$ can be edited to have a separate directory for each domain, make cheat sheets on how to access things, build a small database, make a new tool endpoint for better tool calling, etc. In cases where the environment is highly similar but not identical across tasks, a program can be written to automatically convert a new environment (see \cref{appendix:subsec:auto_convert}). In implementation, we use frontier agents as the research agent, and give it freedom to choose any design choices. The research agent iterates internally over the step-$k$ training tasks and deliver the best $E_k(I)$ it can find at the end of the step. Notably, we do not allow the research agent to modify the harness of $T$ (any changes, though not observed in our experiments, will be discarded at the end of the step). This makes our environment evolution an orthogonal direction to harness optimization, and a good environment can show significant cost reduction even for different agents (\cref{fig:automationbench}).

\begin{algorithm}[tb]
   \caption{Environment Evolution}
   \label{alg:env_evolution}
\begin{algorithmic}
   \STATE {\bfseries Input:} initial environment $E_0(I)$, target agent $T$, research agent $R$, training tasks $X_{\text{train}}$ (no verifiers), validation tasks $X_{\text{val}}$ (with verifiers), number of steps $N$, per-step budget $B$, accuracy tolerance $\delta$
   \STATE Partition $X_{\text{train}}$ into disjoint subsets $X_1, \dots, X_N$
   \STATE $c_0 \gets \mathrm{cost}(T, E_0(I), X_{\text{val}})$
   \STATE $a_0 \gets \mathrm{acc}(T, E_0(I), X_{\text{val}})$
   \FOR{$k=1$ {\bfseries to} $N$}
   \STATE Spawn $R$ in a sandbox with $X_k$, $E_{k-1}(I)$, access to run $T$, and budget $B$
   \STATE $\tilde{E} \gets E_{k-1}(I)$
   \STATE $\tau \gets$ trajectories of $T$ with $\tilde{E}$ on $X_k$
   \REPEAT
   \STATE $R$ analyzes $\tau$ and edits $\tilde{E}$ into a candidate $E'$
   \STATE $\tau' \gets$ trajectories of $T$ with $E'$ on $X_k$
   \IF{cost of $\tau'$ is lower than $\tau$ {\bfseries and} $R$ judges accuracy preserved (proxies / answer consistency)}
   \STATE $\tilde{E} \gets E'$, \; $\tau \gets \tau'$
   \ELSE
   \STATE Discard $E'$
   \ENDIF
   \UNTIL{budget $B$ is exhausted or $R$ terminates}
   \STATE $E_k(I) \gets \tilde{E}$; discard any changes to the harness of $T$
   \STATE $c_k \gets \mathrm{cost}(T, E_k(I), X_{\text{val}})$
   \STATE $a_k \gets \mathrm{acc}(T, E_k(I), X_{\text{val}})$
   \IF{$c_k > c_{k-1}$ {\bfseries or} $a_k < a_{k-1} - \delta$}
   \STATE {\bfseries return} $E_{k-1}(I)$
   \ENDIF
   \ENDFOR
   \STATE {\bfseries return} $E_N(I)$
\end{algorithmic}
\end{algorithm}

\paragraph{Training with No Verifiers} To prevent the research agent from reward hacking during a training step (such as writing down the answers to the training tasks to lower the cost directly), we hide all the verifiers for the tasks used in training. This also mitigates the scarcity of high quality tasks for some stationary environments, as we can synthesize in-distribution queries (without synthesizing the verifiers) for training (see \cref{sec:experiments}). We keep a separate validation set $X_{\text{val}}$ with verifiers and run the target agent with $E_k(I)$ on it at the end of each training step $k$ to decide when to stop the environment evolution. Concretely, if $\mathrm{cost}\bigl(T, E_k(I), X_{\text{val}}\bigr)>\mathrm{cost}\bigl(T, E_{k-1}(I), X_{\text{val}}\bigr)$, or if the pass rate on the validation set becomes significantly worse than step $k-1$, we stop the evolution and use $E_{k-1}$ as the final optimized environment.

\section{Experiments}
\label{sec:experiments}

We evaluate \framework across 5 benchmarks over 3 types of tasks: workspace tasks (AutomationBench \citep{shepard2026automationbench}, Workspace-Bench \citep{tang2026workspacebench10benchmarkingai}), data engineering tasks (Data-Eng-Bench \citep{snowflake2026dataengbench}), and software engineering tasks (Bleve repository tasks as used in \textsc{gskill} \citep{Agrawal_2026} and the SWE Agent Efficiency tasks as used in \textsc{advisor models} \citep{asawa2026trainadvisorsteeringblackbox}). Comparing with the baseline, we observe substantial cost reduction up to 88\%, with higher or similar agent success rate. We also show that across all tested scenarios, the optimized environment can also greatly improve token efficiency and pass rate for different harnesses and different models, including agents stronger than the target agent, demonstrating strong transferability. Notably, \framework can also build on top of harness optimization methods like Meta-Harness and further push the Pareto frontier.

\paragraph{AutomationBench} includes workspace tasks that ask an agent to answer questions or complete work inside an enterprise environment, often covering multiple platforms such as Slack, GitHub, Gmail, Salesforce, etc. From the 600 publicly available tasks, we randomly sample 200 tasks for $X_{\text{test}}$, 96 for $X_{\text{val}}$, and use the rest for the training task pool. Training proceeds with 20 tasks per step and stops at step 5, where results on $X_{\text{val}}$ begin to saturate. We use Claude Code \citep{anthropic2025claudecode} with Claude Opus 4.8 \citep{anthropic2026opus48} as the research agent, and Claude Code with Claude Haiku 4.5 \citep{anthropic2025haiku45} as the target agent.

\begin{figure*}[t]
    \centering
    \includegraphics[width=1\textwidth]{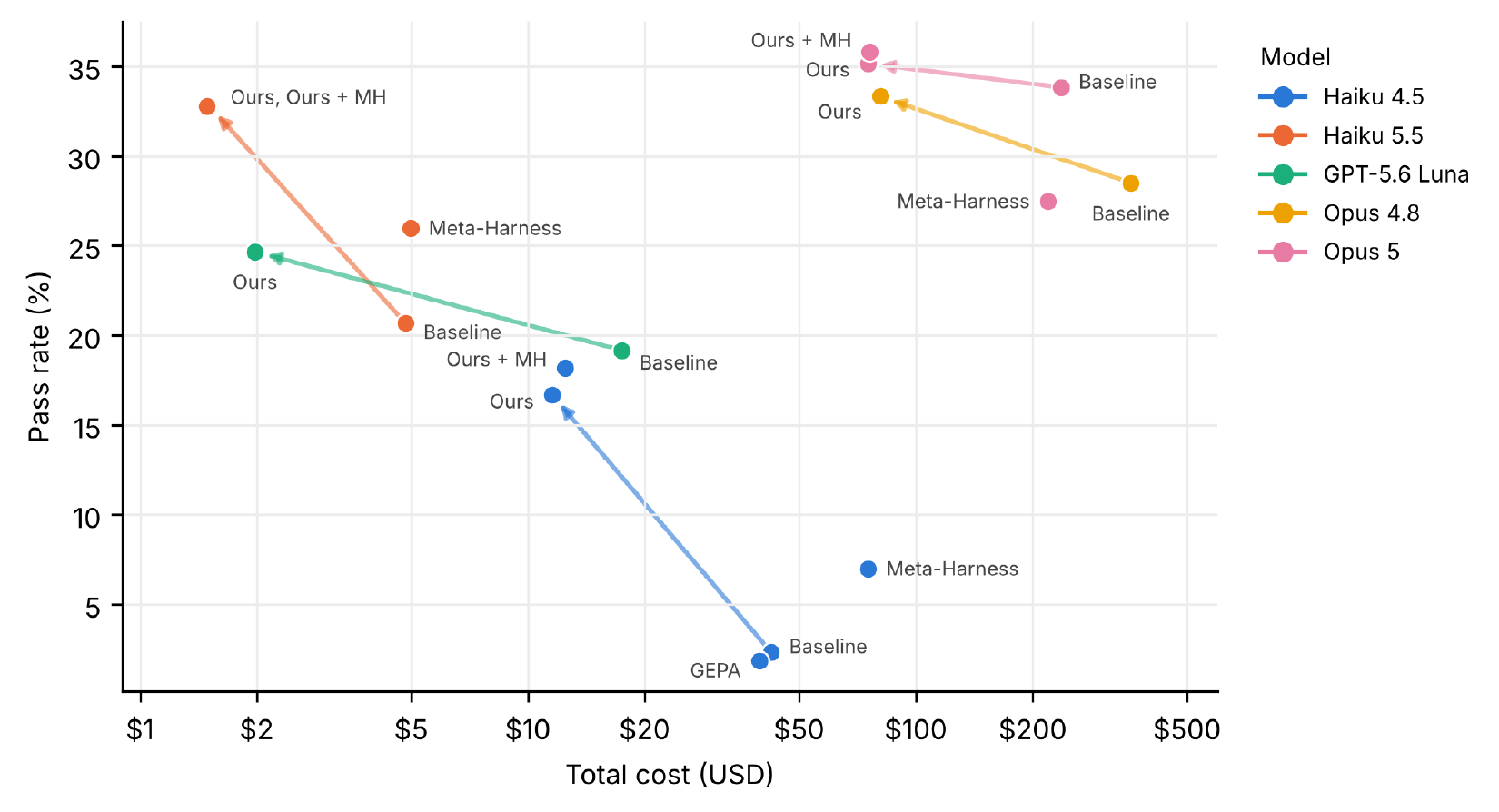}
    \caption{
    Results for AutomationBench. For each agent and each method, we show the pass rate and cost (USD) averaged over 3 independent runs. We use the same color for different methods applied on the same agent, including baseline (original environment), \textsc{GEPA}, Meta-Harness, Ours, and Ours + Meta-Harness. Throughout all tested agents, \framework significantly reduces total cost and improves pass rate, outperforming GEPA and Meta-Harness. Claude Haiku 5.5 with \framework can have success rate similar to Claude Opus 5 with the baseline environment. Additionally, \framework can be applied on top of Meta-Harness, pushing the Pareto frontier for Haiku 4.5 and Opus 5.}
    \label{fig:automationbench}
\end{figure*}

As shown in \cref{fig:automationbench}, we observe substantial improvement of token efficiency and pass rate for the target agent, with a final cost reduction of 72.7\% and pass rate increase of 14.3\%. We show the per-step performance in \cref{appendix:additional_results}. We also run GEPA \citep{agrawal2026gepareflectivepromptevolution} and Meta-Harness \citep{lee2026metaharnessendtoendoptimizationmodel} using the same training task pool, same research agent or proposer, same target agent, and evaluate on the same $X_{\text{test}}$. \framework produces an environment that that achieves significantly better outcomes on both cost and pass rate, even without verifiers for training tasks.

\begin{table*}[t]
\caption{
Detailed results comparing baseline environments and \framework environments on AutomationBench, Workspace-Bench, and Data-Eng-Bench. For AutomationBench, we show results using the official harness released by \citet{shepard2026automationbench}. Agents marked with $\dagger$ were used as the target agent during \framework, while the other agents show transfer results using the environment optimized against the corresponding target agents. Across all agents and benchmarks, \framework significantly reduces cost and improves or maintains pass rate.
}
\label{tab:ws_de_bench}
\vskip 0.15in
\begin{center}
\begin{small}
\begin{tabular}{llccccr}
\toprule
\multirow{2}{*}{Benchmark} & \multirow{2}{*}{Agent} & \multicolumn{2}{c}{Accuracy (\%)} & \multicolumn{2}{c}{Cost (USD)} & \multirow{2}{*}{Cost
red. (\%)} \\
\cmidrule(lr){3-4} \cmidrule(lr){5-6}
& & Base & Evolved & Base & Evolved & \\
\midrule
\multirow{5}{*}{AutomationBench}
& Haiku 4.5$^{\dagger}$ & 2.33  & 16.67 & 42.36  & 11.55 & $-72.73$ \\
& Haiku 5.5             & 20.7  & 32.8  & 4.83   & 1.51  & $-68.74$ \\
& GPT-5.6-luna          & 19.17 & 24.67 & 17.42  & 1.98  & $-88.63$ \\
& Opus 4.8              & 28.50 & 33.33 & 358.13 & 80.90 & $-77.41$ \\
& Opus 5                & 33.83 & 35.17 & 236.18 & 75.26 & $-68.13$ \\
\midrule 
\multirow{3}{*}{Workspace-Bench}
& Codex + GPT-5.4-mini$^{\dagger}$ & 39.84 & 43.12 & 12.55  & 6.57   & $-47.65$ \\
& Codex + GPT-5.6-sol              & 53.98 & 55.05 & 64.59  & 27.45  & $-57.50$ \\
& Claude Code + Opus 5             & 62.27 & 61.14 & 137.41 & 60.64  & $-55.86$ \\
\midrule 
\multirow{3}{*}{Data-Eng-Bench}
& Codex + GPT-5.6-luna$^{\dagger}$ & 50.68 & 53.88 & 16.13  & 7.51   & $-53.44$ \\
& Codex + GPT-5.6-sol              & 62.56 & 61.19 & 182.58 & 79.10  & $-56.68$ \\ 
& Claude Code + Opus 5             & 63.47 & 63.01 & 434.42 & 215.53 & $-50.39$ \\
\bottomrule
\end{tabular}
\end{small}
\end{center}
\vskip -0.1in
\end{table*}

We also examine the transferability of the optimized environment across different harnesses and models, with an average cost reduction of 75.7\% and generally higher pass rates. Notably, \framework with a research agent of Claude Code and Opus 4.8 is able to transfer to Claude Code with Opus 5 \citep{anthropic2026opus5}. Haiku 5.5 \citep{anthropic2026haiku55} with \framework achieves pass rate higher than Opus 4.8 and even close to Opus 5.

Additionally, \framework can be applied on top of Meta-Harness to further improve agent's token efficiency and pass rate. Using the optimized harness for Haiku 4.5, \framework can further push the Pareto frontier for Haiku 4.5. When transferring to Opus 5, although Meta-Harness alone does not show gain on pass rate, \framework on top of Meta-Harness still improves the pass rate with much lower cost.

\paragraph{Workspace-Bench} evaluates agents on workers' personal workspaces, including researchers, engineers, product managers, etc. We use the public set of 100 tasks as used in the official leaderboard and split into 80 tasks for $X_{\text{test}}$ and 20 for $X_{\text{val}}$. Since our training steps do not require verifiers and the number of tasks is sparse for this benchmark, we directly synthesize queries without verifiers with Claude Code and Claude Opus 5 to serve as training tasks. We provide 20 distinct training tasks per step, and use Claude Code with Opus 4.8 as the research agent, Codex \citep{openai2025codex} with GPT-5.4-mini \citep{openai2026gpt54mini} as the target agent. Claude Code with Sonnet 5 is used as the judge agent for evaluation on $X_{\text{test}}$ and $X_{\text{val}}$. We present the results in \cref{tab:ws_de_bench}. We observe 47.65\% in cost reduction and 3.3\% in pass rate improvement for the target agent. On transferability to non-target agents, we evaluate on Codex with GPT-5.6-Sol \citep{openai2026gpt56sol} and Claude Code with Opus 5, and observe an average 56.7\% cost reductions with steady pass rates compared with the baselines. 

\paragraph{Data-Eng-Bench} evaluates agents on dbt data engineering tasks. We use 73 tasks as $X_{\text{test}}$ and 19 as $X_{\text{val}}$. Similarly, due to the scarcity of existing tasks, we synthesized queries without verifiers for training, with 20 distinct tasks per step. We use Codex with GPT-5.6-Sol as the research agent and Codex with GPT-5.6-luna as the target, and reduced cost by 53.44\% without hurting the performance. On non-target agents, we observe an average cost reduction of 53.54\% (see \cref{tab:ws_de_bench}).

\paragraph{Software Engineering Tasks} 

\begin{figure*}[t]
    \centering
    \includegraphics[width=1\textwidth]{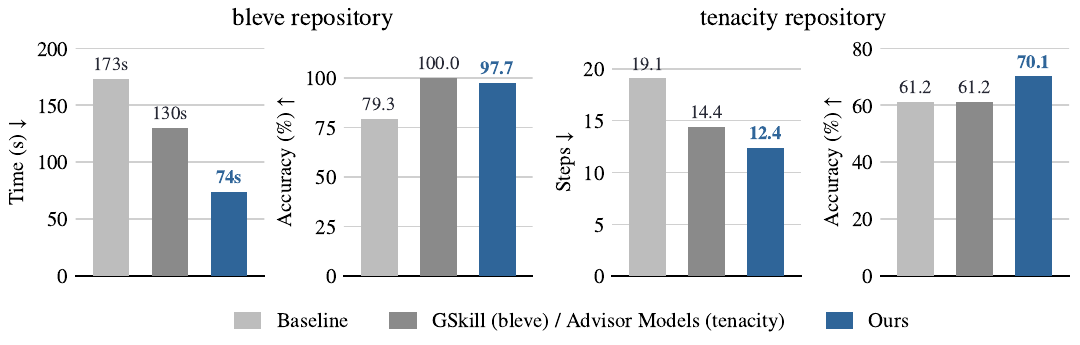}
    \caption{
    Results on two software engineering repositories. \textbf{Left:} on the bleve repository, \framework keeps accuracy close to perfect while using 56.9\% of \textsc{GSkill}’s completion time. \textbf{Right:} on the tenacity repository, \framework reaches 8.9\% higher accuracy than \textsc{Advisor Models} with fewer steps.
    }
    \label{fig:swetasks}
\end{figure*}

We take two benchmarks from previous work that improves token efficiency of agents on SWE-Bench style tasks, where an agent is asked to fix a bug in a given code repository, with all tasks sharing the same codebase. \textsc{gskill} optimizes skills for coding agents using tasks with verifiers from the same codebase and reports higher pass rate and faster completion time on the Bleve search library. \textsc{Advisor models} \citep{asawa2026trainadvisorsteeringblackbox} uses reinforcement learning to train an advisor model that produces advice to the coding agent. On the SWE Agent Efficiency benchmark based on the tenacity library, \textsc{advisor models} maintains the same pass rate as baseline while using significantly fewer steps. We run \framework using the same $X_{\text{train}}$, $X_{\text{test}}$, and $X_{\text{val}}$ as provided by \textsc{gskill} and \textsc{advisor models} and use the same target agent. As shown in \cref{fig:swetasks}, on Bleve repository, \framework remains close to perfect score using only 56.9\% of the completion time of \textsc{gskill}. On the tenacity repository, \framework achieves 8.9\% higher pass rate with fewer steps than \textsc{Advisor Models}. These results show that \framework significantly outperforms other methods to improve agent token efficiency even on software engineering tasks.

\section{Discussion}
\label{sec:discussion}

\subsection{Example Environment Optimizations}

\begin{figure*}[!t]
    \centering
    \includegraphics[width=1\textwidth]{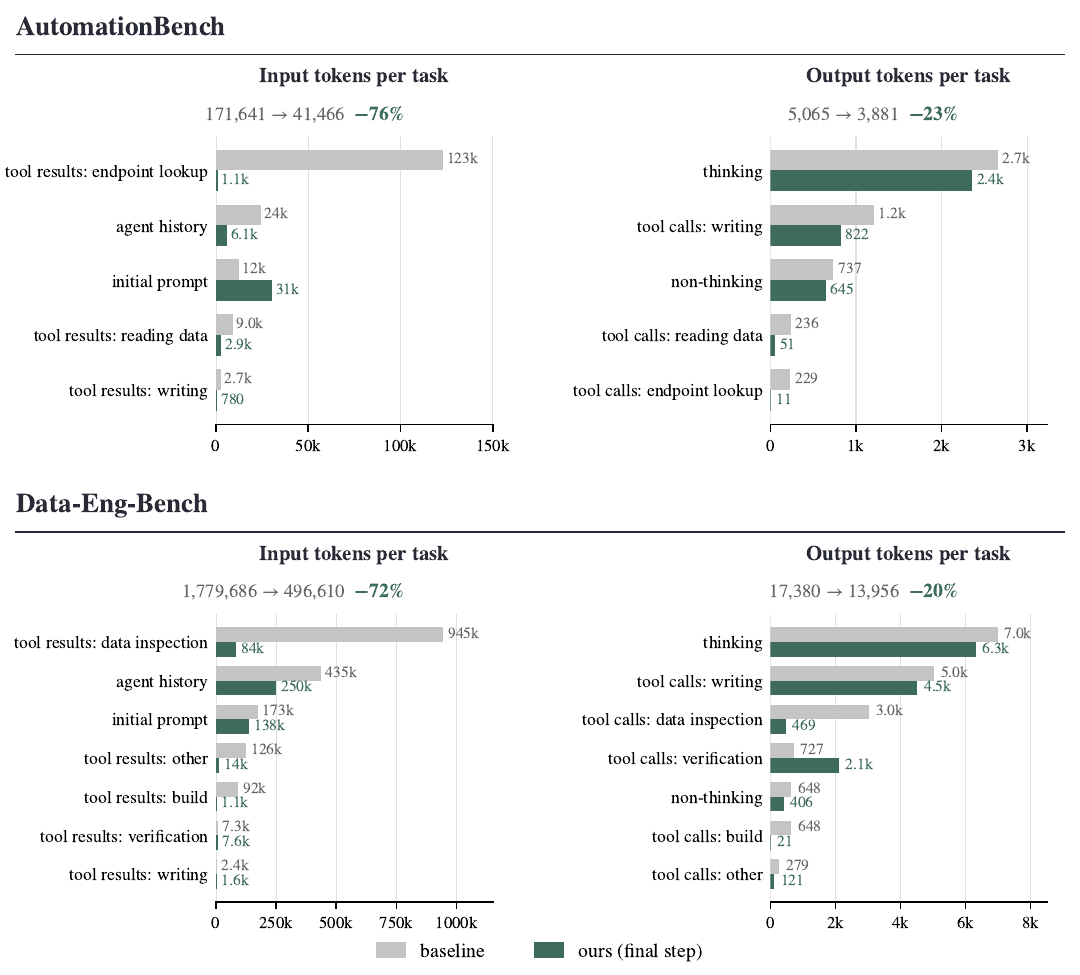}
    \caption{
    Breakdown of token savings by categories on AutomationBench and Data-Eng-Bench. Grey bars show the baseline environment and green bars show the final \framework environment. Most of cost reduction comes from input token savings, mainly tool execution results, such as endpoint lookup on AutomationBench and data inspection on Data-Eng-Bench. Output tokens drop less, around 20\% compared to baseline.
    }
    \label{fig:token_breakdown}
\end{figure*}

\paragraph{Preparing file contents} On AutomationBench, the research agent built a program that directly retrieves Gmail information and presents it as plain text to the target agent, covering the sender, the subject, the message ID, and chunks of the email. The target agent can thus directly read the text instead of spending tokens on massive tool calls. On Workspace-Bench, different formats of files such as \texttt{.xlsx} are converted to \texttt{.txt} files so that the target agent can read the files more easily and do not need to do further parsing.

\paragraph{Making Cheatsheets} Cheatsheets that store important information for future tasks are created during training. For example, a cheatsheet for AutomationBench stores all the API endpoints the agent needs to create Jira issues, post messages on Slack, record a Xero payment, etc., so it does not need to call \texttt{api\_search} or \texttt{api\_fetch} to get the endpoints. For the Bleve repository bug fixes, a cheatsheet was built that stores template commands for complex tool calls, so the agent does not need to compose search commands from scratch.

\begin{table}[t]
\caption{Cost and pass rates for tool-only optimization on AutomationBench and Data-Eng-Bench. Results show that while tool-only optimization can improve pass rates and reduce costs compared to baselines, the full \framework yields significantly greater gains, particularly in cost.}
\label{tab:tools_only}
\vskip 0.15in
\begin{center}
\begin{small}
\begin{tabular}{llcc}
\toprule
Agent & Optimization & Pass (\%) & Cost (USD) \\
\midrule
\multicolumn{4}{l}{\textit{AutomationBench}} \\
\multirow{3}{*}{Haiku 4.5}
    & Baseline       & 2.33  & 42.36 \\
    & Tools-only Opt & 6.33  & 15.75 \\
    & Full \framework       & 16.67 & 11.55 \\
\cmidrule(l){1-4}
\multirow{3}{*}{GPT-5.6-luna}
    & Baseline       & 19.17 & 17.42 \\
    & Tools-only Opt & 21.83 & 2.94  \\
    & Full \framework       & 24.67 & 1.98  \\
\cmidrule(l){1-4}
\multirow{3}{*}{Opus 5}
    & Baseline       & 33.83 & 236.18 \\
    & Tools-only Opt & 35.17 & 100.37  \\
    & Full \framework       & 35.17 & 75.26  \\
\midrule
\multicolumn{4}{l}{\textit{Data-Eng-Bench}} \\
\multirow{3}{*}{Haiku 4.5}
    & Baseline       & 26.94 & 103.27 \\
    & Tools-only Opt & 26.03 & 109.55 \\
    & Full \framework       & 27.85 & 66.36 \\
\cmidrule(l){1-4}
\multirow{3}{*}{GPT-5.6-luna}
    & Baseline       & 50.68 & 16.13 \\
    & Tools-only Opt & 57.08 & 14.60 \\
    & Full \framework       & 53.88 & 7.51  \\
\cmidrule(l){1-4}
\multirow{3}{*}{Opus 5}
    & Baseline       & 63.47 & 434.42 \\
    & Tools-only Opt & 63.93 & 530.58 \\
    & Full \framework       & 63.01 & 215.53 \\
\bottomrule
\end{tabular}
\end{small}
\end{center}
\vskip -0.1in
\end{table}

\begin{table*}[t]
\caption{Training cost (USD) and duration (clock time) for each step of \framework and each benchmark we run. Numbers include the cost and time of the target agent runs launched by the research agent. We also give the total numbers for each benchmark. Overall, the cost and time of running \framework is only a fraction of the cost and time saved.
}
\label{tab:training_cost}
\vskip 0.15in
\begin{center}
\begin{small}
\begin{tabular}{lcccccccccc}
\toprule
 & \multicolumn{2}{c}{AutomationBench} & \multicolumn{2}{c}{Workspace-Bench} & \multicolumn{2}{c}{Data-Eng-Bench} & \multicolumn{2}{c}{Bleve} & \multicolumn{2}{c}{Tenacity} \\
\cmidrule(lr){2-3} \cmidrule(lr){4-5} \cmidrule(lr){6-7} \cmidrule(lr){8-9} \cmidrule(lr){10-11}
 & Cost & Time & Cost & Time & Cost & Time & Cost & Time & Cost & Time \\
\midrule
Step 1 & \$15.33 & 0.5 h & \$71.24 & 1.1 h & \$10.99 & 6.5 h & \$8.11  & 1.8 h & \$11.71 & 1.3 h \\
Step 2 & \$27.37 & 1.2 h & \$86.48 & 1.6 h & \$9.50  & 6.5 h & \$13.30 & 2.6 h & \$7.91  & 0.9 h \\
Step 3 & \$50.48 & 2.2 h & \$28.70 & 2.6 h & \$8.71  & 6.5 h & \$10.25 & 1.8 h & \$6.00  & 0.7 h \\
Step 4 & \$21.37 & 1.0 h & \$36.89 & 1.6 h & \$7.53  & 5.9 h & \$8.12  & 1.2 h & \$7.46  & 0.8 h \\
Step 5 & \$15.54 & 0.7 h & \$55.87 & 5.2 h & ---     & ---   & \$8.59  & 1.5 h & ---     & ---   \\
Step 6 & \$20.01 & 0.9 h & \$60.26 & 3.3 h & ---     & ---   & ---     & ---   & ---     & ---   \\
\midrule
Total  & \$150.10 & 6.6 h & \$339.44 & 15.4 h & \$36.73 & 25.4 h & \$48.37 & 8.9 h & \$33.08 & 3.7 h \\
\bottomrule
\end{tabular}
\end{small}
\end{center}
\vskip -0.1in
\end{table*}

\begin{wrapfigure}{r}{0.5\linewidth}
    \vspace{-1em}
    \centering
    \includegraphics[width=\linewidth]{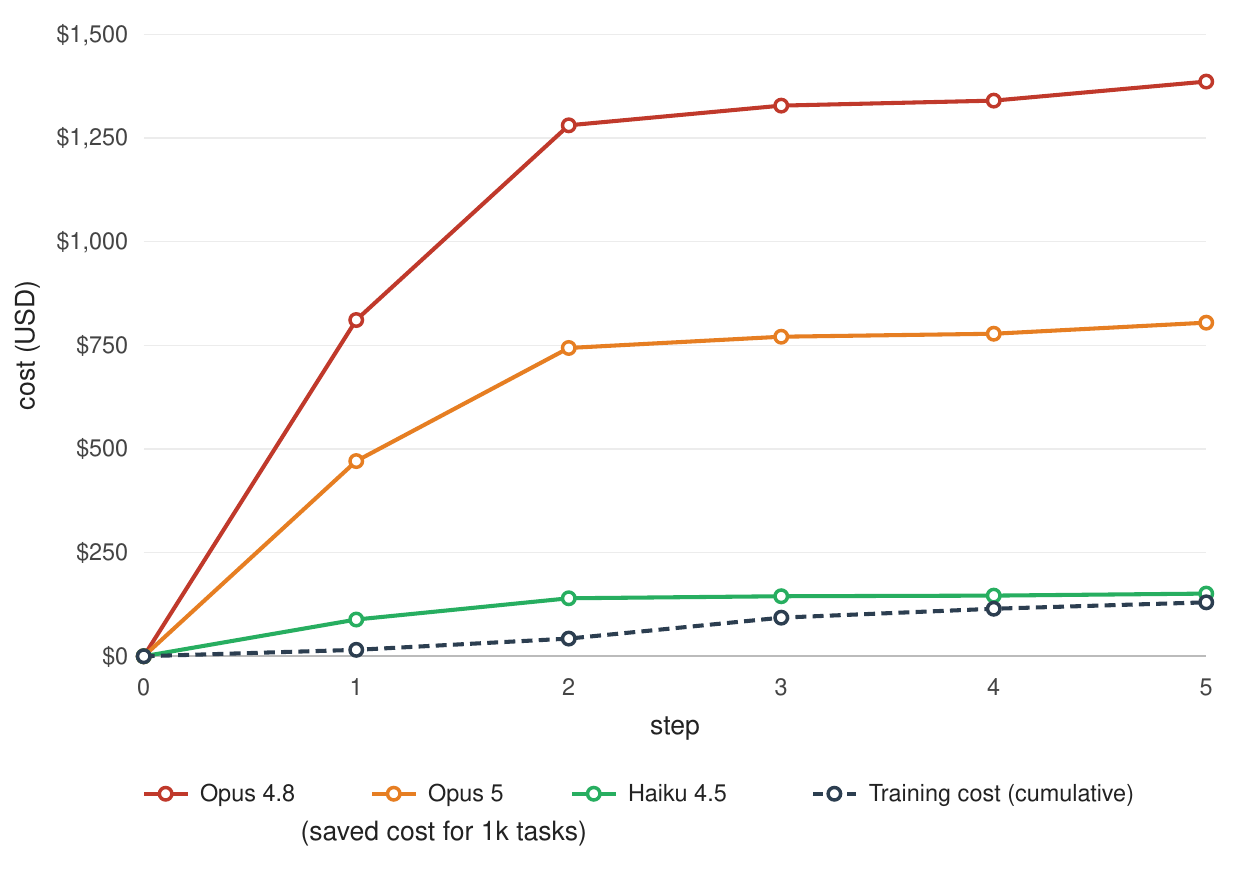}
    \caption{
    Cost savings vs. training cost on AutomationBench. For each agent, solid lines show how much cost each step's \framework environment would save over 1,000 tasks. The dashed line shows the cumulative training cost of \framework over training steps. Savings exceed the training cost for all three agents, by up to about 10× for Opus 4.8.
    }
    \label{fig:automationbench_cost}
    \vspace{-1em}
\end{wrapfigure}

\paragraph{Indexing} For Workspace-Bench, the agent is deployed with massive amount of files simulating an enterprise employee's computer. The research agent thus builds a indexing map as a single text file, that stores the paths, sizes, and types of all files needed, so the can simply search over the indexing file instead of spending tokens and time on the huge filesystem.

\paragraph{Building Database} On Data-Eng-Bench, the research agent built an SQLite database that stores the locations of tables, which files each code script writes to, what are the types of information stored in each table, etc. The target agent can quickly locate the information needed without browsing through thousands of files.

\paragraph{Building Programs that Ranks File Priority} For the Bleve repository, the research agent built a program that takes any bug report and outputs the 8 \texttt{.go} files that are most likely to be impacted based on keyword matching. The target agent can quickly locate the files it potentially need investigate to fix the bugs.

\paragraph{Automated Testing of Edits} For the Tenacity repository, a \texttt{validate\_tenacity.py} is built to automatically fetch the code change made by the target agent, search for relevant test files, run the tests, and return verdicts. This quickly validates candidates edit with no extra token cost and speeds up the whole bug fix workcycle for the target agent.

\subsection{Rejected Environment Optimizations}

We summarize environment optimizations that are discarded during training or after running on $X_{\text{val}}$:

\paragraph{Storing too much information in one file leads to more exploration} For Data-Eng-Bench, the research agent tried to create a big file to store all possible values for all columns in a database. Though this is meant to help the agent retrieve the correct information with less token, putting everything into a big file encouraged the agent to make more tool calls to explore the values, since one tool result reveals to the agent that there is much information it has not explored, which eventually increased the cost.

\paragraph{Building a search tool that returns too many files causes more reading} For Workspace-Bench, a search tool was built to help the agent locate the files that it needs to read for the task. Though the tool successfully reduced token cost for searching the needed files, it often included unnecessary files in the search results, which motivated the agent to read more files, thus increasing the cost.

\paragraph{Adding tests that only fit a subset of the training tasks} In step-2 of \framework on the Bleve repository, a check was added to a test script in the repository that scans for a particular bug fix code snippet and fails the test directly if the snippet is present. This was meant to save the time needed to fix a bug and worked well for $X_2$. However, for $X_3$, some correct bug fixes were rejected by this additional check, and the agent needed to spend much more time to bypass the check.

\subsection{Cost Reduction Analysis}

To illustrate where the cost reduction of \framework comes from, we break down input tokens and output tokens into categories, including different kinds of tool calls, tool execution results (environment feedback), thinking tokens, agent history, etc. We compare the numbers of tokens of each category between the baseline environment and \framework environment, and show results in \cref{fig:token_breakdown}. We observe that the most significant token cost reduction comes from the environment feedback categories for the input tokens. There are also significant savings on the tool call categories of the output token side. These results align with our motivation to optimize the environment for better agent interaction.

\subsection{Ablation: Tool Optimization Only}

During environment optimization, \framework can also optimize ways to access the new environment, including new tools. We study the effectiveness of only using the new tools. On AutomationBench, we test Claude Code with Haiku 4.5 on the original environment $E_0$ with the tools from $E_5$. As presented in \cref{tab:tools_only}, we find that tool optimization alone can lead to improvements on both token efficiency and pass rate, but is significantly lower than optimizing both the tools and the environment. Since tool optimization might be covered by harness engineering, this ablation shows that \framework brings significantly more headroom for optimization in addition to harness engineering.

\subsection{Training Cost and Time}

We present the total training cost (including the cost of target agent runs launched by the research agent) and clock time for all 5 benchmarks in \cref{tab:training_cost}. In \cref{fig:automationbench_cost}, we compare the cumulative training cost with the cost saved for running different agents on unseen tasks, assuming there are 1000 such tasks. We show that for frontier agents deployed in large scale, we can easily save more than 10 times of the cost of running \framework. 

\subsection{Research Agent Behaviors}

Here we present some interesting behaviors we observed from the research agent, including but not limited to failure modes and how behaviors during training can be related to \framework performance on heldout tasks.

\paragraph{Using unreliable verification proxy} Since we do not provide verifiers for training tasks, the research agent for Workspace-Bench occasionally used proxies such as answer length and keyword matching to confirm answer quality during training steps. This often turns out to hurt performance on heldout tasks. After observing validation set, we updated the instruction for research agent to prevent this behavior (see \cref{appendix:prompts}).

\paragraph{Predicting validation performance using research agent behavior} We find that for AutomationBench, cost reduction on validation set is correlated with how much tool call patterns change during each training step as measured in the per-task difference in tool call counts between training runs before and after the step, with Pearson correlation of 0.75. Alternatively, by looking at the fraction of training tasks that received more than 30\% cost reduction in each step, its Pearson correlation with validation set cost reduction is 0.82 for AutomationBench, 0.70 for Workspace-Bench, and 0.56 for Data-Eng-Bench.

\paragraph{Fail to follow the instruction} Despite being told not touch anything related to the model or the harness, the research agent in Workspace-Bench still tried to change the reasoning effort of Codex during its step 6. This change was not effective on heldout tasks, as we only keep the environment directory in the research agent's sandbox and discard the other artifacts. However, such behavior early in one training step could significantly affect the research agent's optimizations during that step, potentially harming the performance.

\paragraph{How far can we go without a validation set} From the tested five benchmarks, we observe that it is generally safe to run \framework for 3 training steps even without running the validation tasks. However, we note that this is only an empirical claim and that the validation set with real verifiers is still important for auditable environment optimization. 

\paragraph{Early flaw in optimization can remain uncovered in many later steps} In Data-Eng-Bench, the research agent introduced a SQL statement splitter in step 1 that parses based on semicolon. But the splitter will fail if the semicolon is in quoted text. This bug in the optimization from step 1 remain uncovered until step 5. This inflates cost through all the 4 steps, as the target agent cannot tell that the splitter corrupted its query, so it attributes the failure to its own model and spends tokens to rebuild it.

\section{Related Work}
\label{sec:related}

\paragraph{Agent Environment Adaptation}

Previous work has explored environment engineering for LLMs or LLM agents for better test-time performance or for training. For test-time deployment, SWE-agent \citep{yang2024sweagent} introduced specialized agent-computer interfaces for software engineering tasks. Subsequently, ALIGN \citep{liu2025agentenvironmentalignmentautomatedinterface} automatically aligns the interface between agents and environments to help with agent understanding of the environment. \citet{xu2026devilinterfaceevaluatingtool,mak2026bashneedempiricalstudy,devrev2026enterprisebench} demonstrate that different tool architectures and interfaces affect agent performance and efficiency. Aegis \citep{song2025aegistaxonomyoptimizationsovercoming} systematically analyzes agent-environment interaction failures and proposes environment optimizations based on the failure taxonomy. SlackBench \citep{li2026slackbenchbenchmarkingagentscollaborative} also found that changing agent's way of access to Slack data from MCP to CLI can improve success rate and lower cost. DocSearch \citep{pmlr-v306-cheng26c} uses test-time feedback to optimize repository documentation for coding agents. Libra \citep{zhao2026libratrainingenvironmentagentic} evolves catalogs as navigable indices in code repositories to facilitate information retrieval. PREPING \citep{choi2026prepingbuildingagentmemory} uses synthetic tasks to construct pre-task procedural memory for agents before they are deployed in target-environmen tasks. META-AGENT \citep{samuel2026studyingsyllabustaskagnosticenvironment} extends the task-agnostic environment preprocessing by adding study artifacts to the environment and shows gains in pass rates while no significant improvement on token efficiency.

Another line of work adapts the environments to facilitate agent training. \textsc{Environment Tuning} \citep{lu2026dontjustfinetuneagent} proposes environment tuning to improve reinforcement learning through corrective feedback, structured curricula, and fine-grained progress rewards. EnvHarness \citep{huang2026envharnessawakeningstaticworlds} introduces programmable environment wrappers that adapt to agent weaknesses and provide improved training signals for reinforcement learning. Agent-World \citep{dong2026agentworldscalingrealworldenvironment} synthesizes diverse interactive environments and verifiable tasks to support agent training.

\paragraph{Harness Optimization}

Harness optimization improves LLM agents by improving the textual wrapper around model parameters, including prompts, tool-use workflows, context management, execution logic, etc. Agent harnesses such as OpenHands \citep{wang2025openhandsopenplatformai} and mini-SWE-agent \citep{yang2024sweagent} provide carefully designed workflows for complex agentic tasks. More recently, \citet{lee2026metaharnessendtoendoptimizationmodel} employs an agentic proposer to automatically optimize harness code by inspecting previous implementations, execution trajectories, and evaluation feedback. Agentic Harness Engineering \citep{lin2026agenticharnessengineeringobservabilitydriven} similarly automates the evolution of coding-agent harnesses, jointly modifying components such as tools, middleware, and memory based on trajectories.

Other work focuses on optimizing individual components of an agent's harness. For prompt optimization, DSPy \citep{khattab2024dspy} introduced a programming framework for optimizing LM pipelines, while GEPA \citep{agrawal2026gepareflectivepromptevolution} uses reflection over execution trajectories to evolve prompts. Combee \citep{li2026combeescalingpromptlearning} further scales prompt optimization to parallel agent executions. Beyond prompts, optimize\_anything \citep{Agrawal_2026} provides a unified framework for optimizing text-based artifacts, including agent architectures and skills. SkillOpt \citep{yang2026skilloptexecutivestrategyselfevolving} iteratively optimizes reusable agent skills through trajectory feedback and validation. Another line of work optimizes agent context and memory. ACE \citep{zhang2026agenticcontextengineeringevolving} evolves structured contextual playbooks through generation, reflection, and curation. Meta Context Engineering \citep{ye2026metacontextengineeringagentic} jointly evolves context-engineering strategies and context artifacts. For agent memory, ReasoningBank \citep{ouyang2026reasoningbankscalingagentselfevolving} distills reusable reasoning strategies from past trajectories and MemEvolve \citep{zhang2025memevolvemetaevolutionagentmemory} jointly optimizes memory contents and memory architectures.

\section{Conclusion}
\label{sec:conclusion}

We formulate optimization over stationary environments as the third direction of agent optimization, alongside harness optimization and parameter tuning. This work introduces \framework, an automated research pipeline for end-to-end environment optimization, and shows that it significantly reduces agent cost while maintaining or improving success rates across five benchmarks. \framework transfers effectively across different models and harnesses, outperforming prior agent optimization methods including GEPA, Meta-Harness, \textsc{gskill}, and \textsc{advisor models}. In addition, \framework can build upon Meta-Harness to push the Pareto frontier even further. Our analysis shows that \framework significantly outperforms tool-only optimization and yields cost savings exceeding 10 times the cost of running \framework itself. Finally, we detail key examples of successful and rejected environment optimizations to serve as a qualitative reference and provide design experience for future research.

\section*{Acknowledgements}
\label{sec:ack}

This project is a collaboration of University of Chicago, UC Berkeley Sky Computing Lab, and Bespoke Labs. This project is partly supported by NSF IFML and Bespoke Labs. Sky Computing Lab is supported by gifts from Accenture, Algorithmic SuperIntelligence Labs, Amazon, AMD, Anyscale, Broadcom, cmpnd, Google, IBM, Intel, Intesa Sanpaolo, Lambda, Lightspeed, Mirendil, NVIDIA, Soma Labs, and VESSL.

\clearpage

\bibliography{custom}
\bibliographystyle{bibstyle}

\clearpage

\appendix

\section{Research Agent Instructions}
\label{appendix:prompts}

We include here the instructions provided to the research agent on the five benchmarks in \cref{sec:experiments}.

\subsection{AutomationBench}

\begin{tcolorbox}[breakable, colback=gray!3, colframe=gray!50, boxrule=0.4pt,
                  arc=2pt, left=5pt, right=5pt, top=4pt, bottom=4pt]
\small\setlength{\parskip}{0.5em}
\textbf{How you work (this session is fully automated)}

No human is watching and you will \textbf{not} get any further instructions --- so do not stop early, do not ask for confirmation, and do not wait. Keep working the whole time.

\begin{itemize}
\item Do many rounds of: \textbf{change the access layer $\rightarrow$ run the target on the training tasks $\rightarrow$ read its cost and what it did $\rightarrow$ keep what helps $\rightarrow$ try the next idea.} Treat ``I made one improvement'' as the start, not the end.
\item When you've genuinely exhausted your ideas \textbf{or reached your spend budget (below)}, leave a short \texttt{REPORT.md} in \texttt{/work}: what you tried, what helped, what didn't, the before/after cost and behavior, and what you'd try next. That report is the finish line.
\end{itemize}

\textbf{You are headless and non-interactive. There is NO one to re-invoke you and NO wakeups/notifications/waiters --- they do not fire here. If you end your turn, the session ends permanently and unfinished work is lost.} Never end your turn to ``wait.'' A benchmark run takes minutes --- run it in the \textbf{foreground} and let it finish, or poll with a blocking loop in the same turn (\texttt{until [ -f /work/out/run.json ]; do sleep 5; done}). Keep your turn alive.

You have \textbf{no web access}. Everything you need is here.

\textbf{Your job}

A fixed AI assistant completes business-workflow tasks across \textasciitilde47 simulated SaaS apps (Gmail, Salesforce, Slack, Google Sheets, QuickBooks, Zendesk, \ldots). It reaches that world \textbf{only through a small tool layer}. Its model and agent are \textbf{fixed --- not yours to touch.} What \emph{is} yours is that \textbf{access layer}: the tools it can call, what they return, \textbf{how the world's data is organized/surfaced before it ever reaches the assistant}, and the instructions that tell it how to work.

\textbf{Make this assistant complete the tasks better AND cheaper by giving it a better access layer} --- push both: \textbf{fewer tokens / tool calls} \emph{and} \textbf{more often actually accomplishing the task.} You are optimizing the \emph{abstraction of the world} the assistant works through --- shaped so the right action is easy and cheap to reach.

There are \textbf{two levers, and the first matters most:}

\begin{enumerate}
\item \textbf{Organize / front-load the world ahead of time.} Each task starts from a fully-populated world. Have the access layer \textbf{pre-fetch it through the base tools at task start, then denormalize / resolve references / lay it out and hand the assistant one compact block} --- so it reads what it needs instead of \emph{discovering} it through many list/get/query calls over noise-padded data. \textbf{This is the bigger lever --- be bold, even radical, here.} Reshape the world's representation; don't just point a better searchlight at the same raw pile.

\textbf{Note --- how ``reshaping the env state'' works here:} unlike a workspace with shared files, every task gets its own fresh world and the assistant has no filesystem --- only these tools. So you can't pre-build one static digest or hand it files. Instead, your access-layer code does the reshaping per task, automatically, at task start.
\item \textbf{The tools + the guide.} Better tools (collapse round-trips, return denormalized results) and an always-on guide that teaches the assistant the efficient, correct pattern.
\end{enumerate}

\textbf{The access layer you edit}

Your access layer is the Python package at \textbf{\texttt{/work/layer}}. \textbf{This is the deliverable} --- edit it in place. \textbf{Read it first:} it may already carry improvements from an earlier pass --- build on those rather than starting over. If it is still the raw baseline (the three plain tools, nothing front-loaded --- a run on it is identical to the stock benchmark), build it up from there.

How the package plugs into the target's run:

\begin{itemize}
\item \textbf{\texttt{LAYER\_TOOLS}} (a list) --- the tools the assistant sees. Each tool takes \textbf{\texttt{base}} as its first argument and the args the assistant passes. \texttt{base} is the only door to the world and is \textbf{hidden} from the assistant's tool spec:
  \begin{itemize}
  \item \texttt{base.search(query, top\_k)} --- discover endpoints (the stock \texttt{api\_search})
  \item \texttt{base.fetch(method, url, params, body)} --- call an endpoint (the stock \texttt{api\_fetch})
  \item \texttt{base.encode(text)} --- base64url for Gmail bodies (the stock \texttt{base64\_encode})
  \end{itemize}
  A tool's body runs \textbf{in your container}; it touches the world only via \texttt{base}. So you can make a tool that chains several \texttt{base.fetch}es and returns one denormalized result.
\item \textbf{\texttt{inject\_context(state, base)}} \emph{(optional --- add it to front-load)} --- called once at task start, before the assistant acts. Pre-fetch the world through \texttt{base}, build a compact overview, and prepend it as a single system message: \texttt{state["prompt"].insert(0, \{"role": "system", "content": ...\})}. This is lever 1 --- and because it's code, not assistant tokens, the pre-fetching is \textbf{cheap in the assistant's cost}.
\end{itemize}

You may add tools, rewrite what they return, add \texttt{inject\_context}, add helper modules / cheatsheets --- anything, as long as the tools reach the world only through \texttt{base}.

\textbf{How you test it --- and how you judge}

You run the target yourself with \textbf{\texttt{evolve-run}} (it runs the assistant on your training tasks through your layer, and writes per-task \textbf{cost} and the full \textbf{trajectory}):

{\fontsize{6}{7.5}\selectfont
\begin{verbatim}
evolve-run \
  --model claude-haiku-4-5-20251001 \
  --base-url https://api.anthropic.com/v1 \
  --api-key-var ANTHROPIC_API_KEY \
  --reasoning-effort high \
  --layer-dir /work/layer \
  --prompts /work/prompts.json \
  --max-concurrent 4 \
  --export-json /work/out/run.json
\end{verbatim}
}

\textbf{There is no answer key and no grade --- you will not see one.} Judge how well the assistant did the way you'd judge any agent: \textbf{read its trajectory and final actions against the task prompt} --- did it find the right records, take the actions the task asked for, and stop? So the loop is: change the layer $\rightarrow$ \texttt{evolve-run} $\rightarrow$ look at \textbf{how much it cost} \emph{and} \textbf{whether it still got the task done} $\rightarrow$ keep what helps, drop what doesn't. \textbf{Always measure on the FULL training set} (every before/after), and run a comparison 2--3 times (cost varies run to run). \texttt{--max-concurrent 4} to iterate faster (don't exceed 4).

You can also \textbf{explore a world by hand} with the same three tools, to understand it before you reshape it (this spends nothing):

{\fontsize{6}{7.5}\selectfont
\begin{verbatim}
probe tasks
  # the worlds you can poke
probe <task> api_search "<query>"
  # discover endpoints in that world
probe <task> api_fetch GET "<url>"
  # read records from that world
\end{verbatim}
}

\textbf{Your budget this step}

You may spend \textbf{up to \$30} running the target this step (sum the \texttt{total\_cost} of your runs). \textbf{This is a ceiling, not a target --- if you reach a good access layer for less, that is better; do not spend just to spend.} Keep a running total and wrap up as you approach \$30. A baseline plus a few candidate comparisons over 20 tasks is most of the cost --- plan experiments so the budget buys the most learning.

\textbf{Haiku 4.5 pricing} (to reckon cost from token counts): \textbf{input \$1 / MTok, output \$5 / MTok}; 5-min cache writes \$1.25/MTok, 1-hour cache writes \$2/MTok, cache hits \& refreshes \$0.10/MTok.

\textbf{The one rule}

The answering model stays \textbf{Claude Haiku 4.5 and must do the real work itself} --- you may not route the task to a different/stronger model. Split by time:

\begin{itemize}
\item \textbf{Offline, before the assistant runs} (\texttt{inject\_context}, helper code) --- wide latitude to pre-fetch and prepare the world so answering becomes easy. But pre-fetching is \textbf{read-only}: do not mutate the world before the assistant acts.
\item \textbf{Online, while it answers} --- only Haiku is in the loop, and the real SaaS actions (\texttt{base.fetch} writes: send email, create/update record, \ldots) \textbf{must actually happen} --- never fake or skip them. The layer only reshapes \emph{reading/orientation}.
\end{itemize}

\textbf{Your tasks}

Your training tasks are the prompts in \textbf{\texttt{/work/prompts.json}} (20 of them) --- \textbf{these are the only tasks in your world.} They are a \textbf{sample of the kinds of tasks this benchmark asks}; the layer you build will be used on \textbf{future, unseen tasks of the same kinds, about different content.} Build something that \textbf{generalizes} --- anything that only fits these exact tasks (hard-coded ids, per-task special-casing) is wasted effort. Keep it general over any world.

\textbf{Deliverable}

Leave your improved layer \textbf{in place at \texttt{/work/layer}} (that is the deliverable) plus \texttt{REPORT.md} (what you changed, the before/after cost and behavior on the full training set, and what you'd try next). Iterate as much as the budget allows; leave the best version you can build.
\end{tcolorbox}

\subsection{Workspace-Bench}

\begin{tcolorbox}[breakable, colback=gray!3, colframe=gray!50, boxrule=0.4pt,
                  arc=2pt, left=5pt, right=5pt, top=4pt, bottom=4pt]
\small\setlength{\parskip}{0.5em}
\textbf{How you work (this session is fully automated)}

No human is watching and you will \textbf{not} get any further instructions --- so do not stop early, do not ask for confirmation, and do not wait. Keep working the whole time. \textbf{Expect this to take hours, not minutes --- a serious optimization here runs for hours, so settle in for a long session and keep iterating until your time or budget is spent.}

\begin{itemize}
\item Do many rounds of: \textbf{change the access layer $\rightarrow$ run the agent on the training tasks $\rightarrow$ read its cost and what it did $\rightarrow$ keep what helps $\rightarrow$ try the next idea.} Treat ``I made one improvement'' as the start, not the end.
\item When you have exhausted your ideas \textbf{or reached your budget (below)}, leave a short \texttt{REPORT.md} in \texttt{/work}: what you tried, what helped, what didn't, and what you'd try next. That report is the finish line.
\item \textbf{Writing \texttt{/work/REPORT.md} ENDS YOUR SESSION IMMEDIATELY AND PERMANENTLY.} The harness watches for that file and stops the run the moment it appears. So never create it early as a draft, placeholder, skeleton, or stub with numbers ``to be filled in'' --- that throws away the rest of your budget and ships a layer you never measured. Write it \textbf{once, at the very end}, only after you have run the treatment on the FULL training set and have the real before/after numbers in hand. Until then keep working notes under any other filename.
\end{itemize}

\textbf{You are headless and non-interactive. There is NO one to re-invoke you and NO wakeups/notifications --- they do not fire here. The session lives ONLY while your turn is active: the moment you end your turn it ends permanently, and any run still going is lost to you.} So never end your turn to ``wait.'' In particular, do \textbf{NOT} launch a benchmark run in the background and stop with something like \emph{``work is proceeding in the background, I'll be re-invoked when it finishes''} --- nothing re-invokes you, so that just throws the run (and your budget) away. A full run can take a long time --- tens of minutes, sometimes hours for many tasks --- and that is \textbf{normal, not stuck}. Stay in this one turn and wait it out: run it in the \textbf{foreground} and let it finish, or if you background it, keep polling it yourself in the same turn until it is done. Only end your turn once you have finished iterating and written \texttt{REPORT.md}.

You have \textbf{no internet access} --- everything you need is in this box.

There is a hard time limit set by the harness; if you reach it the session is killed, so keep moving and leave your \texttt{REPORT.md} before then.

\textbf{Your job}

A fixed agent (codex + gpt-5.4-mini) completes tasks in a \textbf{workspace} --- a real folder tree of files. Its model and harness are \textbf{fixed --- not yours to touch.} What \emph{is} yours is the \textbf{access layer}: how the workspace's data is organized and reached before the agent ever works in it, plus any guidance you leave for how to work in it.

\textbf{This is a big, serious optimization problem --- treat it like a professional re-engineering how an entire company's workspace is accessed, not a quick tweak.} \textbf{Make the agent complete the tasks better by giving it a better access layer.} Be \textbf{bold and creative} --- any kind of access is allowed, and you may transform a workspace \textbf{holistically}: reorganize the data, build a database, add tools or scripts, leave a guide, or change the format entirely. Nothing about the original layout is sacred --- at the extreme you may replace the original files altogether with a completely new representation (say, a single database) if that serves the agent better. How far you go is your call.

\textbf{Above all, do not trade the agent's task performance for cost} --- a change that saves tokens but makes its work worse is a loss, not a win.

You are the engineer responsible for this system's runaway cost. Leadership has been clear: land a \textbf{20\%+ cost reduction} or the project loses its funding --- a seven-figure hit --- and the team is gone. It \textbf{is} achievable --- this is real, not a stretch, so don't settle for a timid tweak: restructure the access layer as aggressively as you need, rebuild it however best serves the agent, and go get the cut. And \textbf{no cheating} --- do not route the work to a different or stronger model, and do not look up benchmark answers online; that gets caught, and the consequences are far worse than missing the target.

\textbf{But it has to generalize.} You are scored on the \textbf{same worlds with different queries you never see.} So transform as freely as you like, but do \textbf{not} tailor the result to the specific tasks in front of you, and however radically you reshape a world, keep everything a different query might still need present and reachable. Winning on the tasks you can see while breaking the ones you can't is the one real failure.

\textbf{The access layer you edit}

Your access layer is at \textbf{\texttt{/work/access/}}, which holds \textbf{one folder per world} --- \texttt{chanpin/}, \texttt{kaifa/}, \texttt{research/}, \texttt{yunying/}, \texttt{houqin/} --- each a real workspace the fixed agent works in. Your training tasks span these worlds. \textbf{This is the deliverable} --- edit the world folders in place. \textbf{Study the current state of each world first, then make it better.} Treat this as the access layer that will actually be used --- do your most complete, ambitious work now.

Each task's input files are already inside its world (the normal workspace files) --- there is no separate inputs folder to manage.

\textbf{How you test it --- and how you judge}

Run the agent yourself with the \textbf{same command the benchmark uses} --- it runs codex + gpt-5.4-mini on your training tasks against your current \texttt{/work/access}, each in its own container:

{\fontsize{6}{7.5}\selectfont
\begin{verbatim}
docker compose -f /work/wb-compose.yaml \
  run --rm workspace-bench bash \
  /workspace/Workspace-Bench/evaluation/docker/\
run-benchmark.sh \
  --harness codex --model gpt-5.4-mini \
  --model-name GPT-5.4-mini --env-prefix GPT54MINI \
  --dataset full --run-name "run_$(date +%s)"
\end{verbatim}
}

\begin{itemize}
\item \textbf{Run it exactly as shown --- do NOT prepend \texttt{export WB\_RUN\_DIR=/work/...} or \texttt{WB\_ACCESS\_DIR=/work/...}.} Those \texttt{/work} paths exist only inside your shell, not on the host where the container is actually launched, so overriding them makes the run fail with \texttt{run-benchmark.sh: No such file or directory}. The correct paths are already set for you.
\item \textbf{Every run must use a distinct \texttt{--run-name}.} The runner \textbf{skips any task whose output already exists}, so a reused name silently reports the \emph{previous} run's results and ruins your before/after. The \texttt{run\_\$(date +\%s)} above auto-generates a fresh name each time; do not hard-code a fixed string.
\item A run's results land under \texttt{/work/run/output/\allowbreak Codex--GPT-5.4-mini--<run-name>/\allowbreak <task>/} (your newest such directory is the latest run): \texttt{agent.log} (trajectory), \texttt{agent.json} (cost), and codex's outputs.
\item \textbf{Establish a real baseline first: run the ENTIRE training set (every task), not a sample --- and run it more than once, because the weak agent's cost and behavior swing run-to-run.} Use a full, repeated run for every before/after you claim. \texttt{--task-limit} is only for a quick sanity check while building --- never for a baseline or a reported result. The budget is generous; do not skimp on measurement.
\item \textbf{There is no answer key and no grade.} Judge how well the agent did the way you'd judge any agent: read its \textbf{trajectory} and the \textbf{files it produced} against the task prompt --- did it find the right files, do what the task asked, produce a sound result? --- and read its \textbf{cost}. Keep what helps, revert what doesn't. Repeat until your budget is spent.
\item \textbf{Optimize for cost --- the dollar figure above --- but without significantly hurting how well the agent does the task; small differences don't matter, so do not overshoot chasing them.} Do \textbf{not} rank designs by output length/size or any proxy; you have no answer key, so those are meaningless. Keep only what lowers real cost while the agent still does the task as well.
\end{itemize}

\textbf{Rigor}

\textbf{Every run and every comparison uses the FULL training set --- every task, every time --- on both axes: cost AND how well the agent did the task.} Never sample, never \texttt{--task-limit}, never eyeball a hand-picked few. A number measured on a subset is not a result --- it is exactly how a real regression slips through.

\textbf{Your budget this step}

You may spend up to \textbf{\$100} of gpt-5.4-mini usage this step. Each task's \texttt{agent.json} reports \texttt{promptTokens} and \texttt{completionTokens}; reckon a run's cost as \texttt{input \$0.75/1M, cached input \$0.075/1M, output \$4.50/1M}, summed over its tasks, and keep a running total across runs. This is a \textbf{ceiling, not a target}: a good access layer for less is better, so do not spend just to spend. A baseline plus a few candidate comparisons over your tasks is most of the budget; plan experiments so each one buys the most learning.

\textbf{gpt-5.4-mini pricing (official):} input \textbf{\$0.75 / 1M tokens}, cached input \textbf{\$0.075 / 1M}, output \textbf{\$4.50 / 1M}.

\textbf{The one rule}

The agent doing the work stays \textbf{codex + gpt-5.4-mini and must do the real work itself} --- you may not do the task for it or precompute its answers. Reshape \emph{how the data is organized}, not \emph{what the answer is}. And please don't route the work to a different model --- the fixed agent being tested should do the tasks itself, not hand them off to a stronger or external one.

\textbf{Your tasks}

Your training tasks are in \textbf{\texttt{/work/tasks/}} --- the only tasks given to you. They span the 5 worlds and are a \textbf{sample of the kinds of tasks these workspaces get}; the access layers you build will be used on \textbf{future, unseen tasks of the same kinds, about different content.} Build something that \textbf{generalizes.}

\textbf{Deliverable}

Leave your improved workspace \textbf{in place at \texttt{/work/access/}} plus \texttt{REPORT.md} (what you changed, the before/after on your tasks, what you'd try next).

\textbf{Keep a record of every version you build (its cost and how well the agent did the tasks).} The deliverable in \texttt{/work/access/} must be \textbf{the best version you built --- the lowest cost that does not significantly harm how well the agent completes the tasks --- not necessarily your latest.} Do not keep iterating and then ship a stripped-down ``safe'' fallback that threw away a better earlier version; pick the genuine best of what you built and leave that in \texttt{/work/access/}. Iterate as much as the budget allows.
\end{tcolorbox}

\subsection{Data-Eng-Bench}

\begin{tcolorbox}[breakable, colback=gray!3, colframe=gray!50, boxrule=0.4pt,arc=2pt, left=5pt, right=5pt, top=4pt, bottom=4pt]
\small\setlength{\parskip}{0.5em}
\textbf{How you work (this session is fully automated)}

No human is watching and you will \textbf{not} get any further instructions --- so do not stop early, do not ask for confirmation, and do not wait. Keep working the whole time. \textbf{Expect this to take hours, not minutes.} A serious optimisation here runs for hours, so settle in and keep iterating until your time or your budget is spent.

\begin{itemize}
\item Do many rounds of: \textbf{change the access layer $\rightarrow$ rebuild the image $\rightarrow$ run the target on the training tasks $\rightarrow$ read its cost and what it did $\rightarrow$ keep what helps $\rightarrow$ try the next idea.} Treat ``I made one improvement'' as the start, not the end.
\item When you have exhausted your ideas \textbf{or reached your budget}, leave a short \texttt{REPORT.md} in \texttt{<STEPDIR>}: what you tried, what helped, what didn't, and what you would try next. That report is the finish line.
\end{itemize}

\textbf{You are headless and non-interactive. Nothing re-invokes you and there are NO wakeups or notifications. The session lives ONLY while your turn is active: the moment you end your turn it ends permanently, and any run still going is lost to you.} So never end your turn to ``wait''. Do not launch a benchmark run in the background and stop with something like \emph{``the run is proceeding, I'll be re-invoked when it finishes''} --- nothing re-invokes you, and that throws the run and your budget away. A full run over the training tasks takes tens of minutes and sometimes hours. That is \textbf{normal, not stuck}. Stay in this one turn: run it in the foreground and let it finish, or if you background it, keep polling it yourself in
the same turn until it is done.

There is a hard time limit of about \textbf{7.0 hours}. When it expires the session is killed wherever it happens to be, so leave your \texttt{REPORT.md} before then.

\textbf{Internet}

You do not need the internet and your browsing tool is switched off. Everything you need is in this box: the warehouse, the dbt project, the training tickets, and the prices below. Work from those.

\textbf{Do not look up this benchmark, its tasks, its repository, or any solutions, anywhere, by any means, including from a shell.} The tickets you are given are a sample of a public benchmark whose reference solutions are published. Finding them would not help you --- you are not scored on task answers --- and it would invalidate the entire experiment. This is the one absolute line.

\textbf{Your job}

A fixed agent --- \textbf{codex + gpt-5.6-luna} --- does data engineering tickets against a retail warehouse. It writes and repairs dbt models. Its model, harness and prompt are \textbf{fixed and are not yours to touch.}

What \emph{is} yours is the \textbf{access layer}: what the agent finds already present in its container, how the warehouse and the dbt project are organised, indexed and described before it starts, and any guidance you leave for working in them.

\textbf{Make the fixed target complete its tasks more cheaply by improving that access layer.} Be bold. Build indexes, write catalogues, precompute summaries, add scripts, leave a guide, restructure what is there. How far you go is your call. Nothing about the existing layout is fixed: you may reorganise, relocate or suppress what is already there, not only add to it.

\textbf{But it has to generalise.} You are scored on the \textbf{same warehouse with different tickets you never see}. So build for the shape of the work, not for the tickets in front of you. Winning on the tasks you can see while breaking the ones you cannot is the one real failure.

\textbf{The access layer you edit}

It is the directory \textbf{\texttt{<STEPDIR>/\allowbreak bench/\allowbreak access-image/\allowbreak access/}}.

Everything in it is copied into every task container at \textbf{\texttt{/app/access/}} when the image is built. The target agent's working directory is \texttt{/app}, so \texttt{/app/access/} shows up the first time it lists the directory. That is your channel to it: there is no other.

\textbf{This is the deliverable.} Edit it in place.

\textbf{What the target actually faces}

Read one of the training tickets in \texttt{<STEPDIR>/tasks/} before you do anything else. The dbt project and warehouse are large: 19 schemas, several hundred source tables, and thousands of model files. Every ticket asks the agent to write or repair a dbt model, and every ticket puts the agent through the same preliminaries before it can start.

Find out what those preliminaries cost by reading a real trajectory, not by guessing.

\textbf{How you test it}

Rebuild the image, then run the target the same way the benchmark does:

{\fontsize{6}{7.5}\selectfont
\begin{verbatim}
cd <STEPDIR>/bench && docker build access-image/ \
-t data-eng-bench-base:<STEP> \
&& harbor run --config configs/data-eng-bench-duckdb.codex.yaml \
     --path <STEPDIR>/tasks --n-attempts 1 \
     --disable-verification --cpus ignore --memory ignore \
     --ae CODEX_AUTH_JSON_PATH=/root/.codex/auth.json \
     --allow-agent-host chatgpt.com --allow-agent-host auth.openai.com \
     --allow-agent-host api.openai.com
\end{verbatim}
}

The rebuild is what installs your access layer. If you edit \texttt{access/} and do not rebuild, you are measuring the previous version and will draw the wrong conclusion.

\texttt{--disable-verification} is required and must stay. These training tickets ship no grader on purpose, and without that flag harbor discards every one of them and reports that the dataset is empty.

\texttt{--cpus ignore --memory ignore} is also required and must stay. This box runs its own docker engine, and that engine cannot apply the per-container CPU and memory limits the tasks declare --- every container fails to start with a cgroup error. Ignoring the limits is what lets them run at all.

Results land under \texttt{<STEPDIR>/\allowbreak bench/\allowbreak jobs/\allowbreak <timestamp>/}. Per trial you get \texttt{result.json} with \texttt{cost\_usd}
and the token counts, \texttt{agent/\allowbreak codex.txt} with the full trajectory, and \texttt{trial.log}. Read trajectories, not just totals: the totals tell
you that something got cheaper, the trajectories tell you why, and only the second kind of knowledge transfers to the next idea.

\begin{itemize}
\item \textbf{Every run and every comparison uses the full training set} --- every task, every time, for cost and for whether the work still got done. A number measured on a subset is not a result. Comparing a hand-picked few is exactly how a regression ships.
\item \textbf{Run the baseline more than once before you trust it.} This target's cost swings run to run. Judge the typical case, not one lucky run.
\item \textbf{A grep is not a check.} Reading your own index tells you what you wrote, not what the target did with it. Open the trajectory and confirm the target opened what you left.
\item \textbf{Verify every claim on an actual run before you build on it.} If something is unverified, say so and go test it.
\end{itemize}

\textbf{How you judge}

\textbf{There is no answer key and no grade.} Nothing here tells you whether the target got the ticket right, and you cannot find out. Judge the way you would judge any engineer's work: read the trajectory and the SQL it produced against what the ticket asked, and read the cost.

\textbf{Optimise for cost. Do not trade the work getting done for a cheaper run.} If you are confident a change also makes the target more likely to succeed, so much the better, but the floor is that it must not get worse.

\textbf{[IMPORTANT NOTE]} We do not want an optimisation that lowers the cost but also degrades performance, so you are discouraged from any optimisation that encourages the target agent to guess, to skip verification, or to do things without checking.

\textbf{Your budget this step}

You may spend up to \textbf{\$40} of gpt-5.6-luna usage this step. Harbor reports \texttt{cost\_usd} per trial in \texttt{result.json}; sum it across your runs and keep a running total.

Official rates, per 1M tokens:

\begin{center}\small
\begin{tabular}{llll}
model & input & cached input & output \\
\hline
gpt-5.6-luna & \$0.20 & \$0.02 & \$1.20 \\
gpt-5.6-sol & \$4.00 & \$0.40 & \$20.00 \\
\end{tabular}
\end{center}

Long-context variants are charged at higher rates. Most of your input is cached, so the cached rate is the one that dominates a repeated run --- a change that raises the uncached share of the prompt can cost more than it saves even when it shortens the trajectory.

This is a \textbf{ceiling, not a target}: a good access layer for less is better, so do not spend just to spend. A repeated baseline plus a few candidate comparisons is most of the budget. Plan each run to buy the most learning.

Both agents run on a subscription, so these dollars are a computed equivalent rather than a bill. Track them anyway --- they are the objective.

\textbf{What good looks like}

This is a cost-control project and the number matters.

A reduction of at least \textbf{20\%} is achievable here. Treat that as the floor, not the goal. The same method has been measured at 35\% to 60\% on comparable benchmarks, by people working with no more information than you have, against warehouses no better organised than this one.

So if you find yourself concluding after a few rounds that the layout is already close to optimal and there is nothing left worth doing, you are wrong and you have stopped too early. That conclusion has been reached prematurely on every benchmark this method has been run on, and it has been wrong every time. Go and read another trajectory. The target is repeating work on every single ticket that it should not have to repeat once.

Two failure modes to avoid, in order of how often they happen:

\begin{itemize}
\item Stopping at the first idea that works. One good change is a start. The measured results above came from ten or more rounds, not from one.
\item Polishing a change that already works instead of looking for the next unexploited source of waste. When a round produces only a marginal gain, that is a signal to go back to the trajectories and find something different, not to tune what you have.
\end{itemize}

\textbf{The one rule}

The agent doing the work stays \textbf{codex + gpt-5.6-luna and must do the real work itself}. You may not do its tickets for it, pre-write the dbt models it is asked to write, or leave answers where it will find them. Reshape \textbf{how the warehouse and the project are organised and reached}, never \textbf{what the answer is}.

\textbf{Your tasks}

Your training tickets are in \textbf{\texttt{<STEPDIR>/tasks/}} and are the only tickets you get. They are a \textbf{sample of the kinds of work this warehouse gets}; the access layer you build will be used on \textbf{future, unseen tickets of the same kinds, against the same warehouse}. Build something that generalises.

\textbf{Deliverable}

Leave your improved access layer in place at \textbf{\texttt{<STEPDIR>/bench/access-image/access/}}, plus \textbf{\texttt{<STEPDIR>/REPORT.md}} covering what you changed, the before and after cost on your training tasks, and what you would try next.

\textbf{Deliver the best version you built, not whatever state your last experiment happened to leave behind.} If a change made things worse, revert it. Keep snapshots of promising versions as you go so you can restore your best one before you finish.
\end{tcolorbox}

\subsection{Bleve}

\begin{tcolorbox}[breakable, colback=gray!3, colframe=gray!50, boxrule=0.4pt,
                  arc=2pt, left=5pt, right=5pt, top=4pt, bottom=4pt]
\small\setlength{\parskip}{0.5em}
\textbf{Your job}

You are the engineer responsible for the runaway cost of an automated bug-fixing system. A \textbf{fixed} agent --- \textbf{Claude Code running Claude Haiku 4.5}, which you cannot change --- fixes bugs in the \texttt{blevesearch/bleve} Go codebase, and it is far too expensive: it burns too many tokens and steps on every bug. Leadership has been clear: \textbf{cut that cost substantially --- a 20\%+ reduction --- or this project loses its funding and the team is gone.} It \textbf{is} achievable; this is real, not a stretch. And 20\% is a floor, not a finish line --- the more you cut, the better; do not stop the moment you clear it.

You should optimize the cost. While we don't put accuracy as your primary objective, it would be nice if you can boost accuracy too, but at least do not make accuracy worse by your optimizations.

What you reshape is the agent's \textbf{environment} --- the entire world the agent works in. Not just ``tools'': it's what the agent has available \emph{and} how the codebase is organized and represented to it. You decide how that world is laid out, indexed, described, and equipped for the agent. Reshaping the environment so the fixed agent works far more cheaply is the whole job.

\textbf{Be bold.} You have wide latitude --- rethink and reshape the environment as aggressively as you need. Timid additions won't get a 20\% cut. This is a real, hard engineering problem; treat it like one. And what you build must \textbf{generalize} to unseen future tasks as well.

\textbf{Make the agent faster, never lazier.} There are two ways to make a step cheaper: help the agent reach the answer sooner, or forbid it from doing the work at all. Only the first is optimization; the second is sabotage that happens to score well on your budget. Every saving you keep must come from better information --- a sharper map, a faster command, a clearer starting point --- never from a ceiling on effort, a ban on a kind of check, or an instruction to stop while the agent is still unsure. You cannot see the grade, so a rule that removes work always looks like a pure win to you: the cost drops and nothing visibly breaks. It breaks on the bugs you never ran. Before you keep any rule, ask what it costs the agent on a task where its first guess was wrong --- if the honest answer is ``it can no longer recover,'' delete it, however much it saved.

\textbf{What you have (in /work)}

\begin{itemize}
\item \texttt{broken/<id>/repo/} --- the bleve repo with one real bug already in it (its tests removed).
\item \texttt{broken/<id>/problem.txt} --- the bug report the agent is handed for that repo.
\item \texttt{layer/} --- everything you build. It is added into the agent's copy of the repo before the agent starts, becoming part of its environment. The agent is \textbf{Claude Code}, which automatically loads a file named \texttt{CLAUDE.md} at the repo root on every turn --- so put any startup guidance you want it to always see in \texttt{layer/CLAUDE.md} (named exactly that). It does \textbf{not} read \texttt{AGENTS.md}; don't use that name.
\end{itemize}

\textbf{How to work}

To see how the agent does on a task with your current environment, run:

{\fontsize{6}{7.5}\selectfont
\begin{verbatim}
python3 /work/box_run.py /work/broken/<id> \
  --layer /work/layer --out /work/evolve_log/<name>
\end{verbatim}
}

It runs the fixed agent on that broken repo inside your environment and writes the agent's full trajectory (\texttt{trajectory.jsonl}) and its code change (\texttt{fix.diff}), and prints the run's \textbf{cost} (\texttt{cost\_usd}) and token use. \textbf{Read the trajectory} --- it is your only signal. See where the agent wastes tokens and steps, and reshape the environment to remove that waste.

\textbf{Also watch out for correctness --- do not let cost reduction lead to any decrease in correctness.} To judge accuracy, don't rely on cost, or on whether the fix's diff matches the baseline run --- a matching diff only shows the agent did the same thing twice, not that it's correct. Check correctness yourself.

\textbf{You must do your optimization and tests on all the tasks you have, not just a handful of them.}

\textbf{You must do your optimization and tests on all the tasks you have, not just a handful of them.}

\textbf{You must do your optimization and tests on all the tasks you have, not just a handful of them.}

\texttt{box\_run} prints each run's cost directly as \texttt{cost\_usd} (Claude reports it, cache-aware). That cost is exactly what you are driving down --- measure it per task and track your total.

\textbf{Your budget for this step is \$50} of target-run spend. Keep a running total across your \texttt{box\_run} calls and stop before you exceed it.

\textbf{You are headless and non-interactive. There is NO one to re-invoke you and NO wakeups/notifications --- they do not fire here. The session lives ONLY while your turn is active: the moment you end your turn it ends permanently, and any run still going is lost to you.} So never end your turn to ``wait.'' In particular, do \textbf{NOT} launch a \texttt{box\_run} in the background and stop with something like \emph{``work is proceeding in the background, I'll be re-invoked when it finishes''} --- nothing re-invokes you, so that just throws the run (and your budget) away. Running across all the tasks can take a long time --- tens of minutes, sometimes hours --- and that is \textbf{normal, not stuck}. Stay in this one turn and wait it out: run in the \textbf{foreground} and let it finish, or if you background it, keep polling it yourself in the same turn until it is done. Only end your turn once you have finished iterating and written \texttt{REPORT.md}.

\textbf{Rules}

\begin{itemize}
\item \textbf{Do not lower how often the agent fixes bugs.} Cheaper is worthless if it stops fixing them --- cost down, fixes intact.
\item \textbf{Any validation helper you build must fail when the tests it runs fail.} Never downgrade a failing test run to a weaker check such as compile-only, never cache or suppress a previous pass, and never print a success message on a path where tests did not actually pass.
\item \textbf{Never cap how many times the agent may look.} Do not write fixed limits on searching, reading, or validating --- no ``exactly one search'', ``validate once then stop'', ``do not check twice''. A bug report names where the failure \emph{surfaced}, which is often not where the bug \emph{is}, so the agent must stay free to follow the trail to the real cause. Speed must come from pointing the agent at the right place, never from forbidding it a second look.
\item \textbf{You only ADD to the environment; the repo's existing source stays exactly as it is.} Every existing file, identifier, function signature, and error message must remain byte-for-byte unchanged --- do not rename, move, rewrite, overwrite, or delete existing code. (The bug and its hidden tests are pinned to that exact code; altering any of it corrupts the task.) Build your environment \emph{alongside} the code, never by editing it.
\item \textbf{No cheating.} Don't swap in a different or stronger agent/model. Don't hardcode answers to these specific bugs --- you're judged on bugs you've never seen, so only \emph{general} improvements count. Cheating gets caught, and the consequences are worse than missing the target.
\item You never see the tests or the grade --- judge only from the agent's trajectories.
\end{itemize}

\textbf{When done}

Write \texttt{/work/REPORT.md}: what you reshaped, the waste you saw in the agent, and the cost effect you observed.
\end{tcolorbox}

\subsection{Tenacity}

\begin{tcolorbox}[breakable, colback=gray!3, colframe=gray!50, boxrule=0.4pt,
                  arc=2pt, left=5pt, right=5pt, top=4pt, bottom=4pt]
\small\setlength{\parskip}{0.5em}
\textbf{Your job}

You are the engineer responsible for how many steps an automated bug-fixing system burns. A \textbf{fixed} agent --- \textbf{mini-swe-agent driving Gemini 2.5 Flash}, which you cannot change --- fixes bugs in the \texttt{jd/tenacity} Python repository, and it takes far too many steps to do it. Each step is one shell command --- the agent writes its reasoning, then exactly one command, which may chain with \texttt{\&\&} or \texttt{||}. It is cut off at \textbf{40 steps}; a task that hits the cap counts as unresolved.

Leadership has been clear: \textbf{cut the average step count substantially --- a 20\%+ reduction --- or this project loses its funding and the team is gone.} It \textbf{is} achievable; this is real, not a stretch. And 20\% is a floor, not a finish line --- the more you cut, the better; do not stop the moment you clear it.

Steps are the number you are judged on. While we don't put the resolve rate as your primary objective, it would be nice if you can raise it too, but at least do not make it worse by your optimizations.

What you reshape is the agent's \textbf{environment} --- the entire world the agent works in. Not just ``tools'': it's what the agent has available \emph{and} how the codebase is organized and represented to it. You decide how that world is laid out, indexed, described, and equipped for the agent. Reshaping the environment so the fixed agent works in far fewer steps is the whole job.

\textbf{Be bold.} You have wide latitude --- rethink and reshape the environment as aggressively as you need. Timid additions won't get a 20\% cut. This is a real, hard engineering problem; treat it like one.

\textbf{How you are actually graded --- read this twice}

\textbf{Your layer is scored on a set of bugs you will never see.} The tasks in \texttt{tasks/ids.txt} are a development sample and nothing more. When you are finished, your layer is taken away and run against a held-out set of different bugs in this same repository --- different functions, different modules, different symptoms. Your score is the step count \emph{there}. Nothing you achieve on the tasks in front of you counts for anything by itself.

So the question to ask of every line you write is not ``does this help on my sample?'' but \textbf{``will this still help on a bug I have not seen?''}

This is not a ban on writing things down. A cheat-sheet, a source map, a set of invariants, a triage script --- all of these are exactly the right shape, \emph{provided they are about the codebase rather than about the ten bugs you happened to be given}. ``\texttt{tenacity/wait.py} holds every wait strategy'' is about the codebase and will still be true next week. ``Do not add 1 to the ordinal in \texttt{after\_log}'' is about one bug you saw; on the held-out set it is dead weight at best, and actively misleading at worst, because the next bug in that file will be something else entirely.

The honest test for any line: \textbf{would you have written it if you had never seen that particular task?} If no, it will not earn you anything on the held-out set --- delete it and spend the space on something that will.

\textbf{Make the agent faster, never lazier.} There are two ways to make a step count smaller: help the agent reach the answer sooner, or forbid it from doing the work at all. Only the first is optimization; the second is sabotage that happens to score well on your metric. Every saving you keep must come from better information --- a sharper map, a faster command, a clearer starting point --- never from a ceiling on effort, a ban on a kind of check, or an instruction to stop while the agent is still unsure. You cannot see the grade, so a rule that removes work always looks like a pure win to you: the step count drops and nothing visibly breaks. It breaks on the bugs you never ran. Before you keep any rule, ask what it costs the agent on a task where its first guess was wrong --- if the honest answer is ``it can no longer recover,'' delete it, however much it saved.

\textbf{What you have (in \texttt{/work})}

\begin{itemize}
\item \texttt{tasks/} --- the training task ids for this step, one per line in \texttt{ids.txt}.
\item \texttt{layer/} --- everything you build. Its contents are placed into the agent's checkout of the repository (at \texttt{/testbed}) before the agent starts, becoming part of its environment.

\textbf{\texttt{layer/AGENTS.md} is mandatory and is the agent's only entry point.} The target is mini-swe-agent, which auto-loads nothing --- it does not read any file unless told to. The one thing the agent is told is a single fixed line appended to the task: that maintainer notes and tooling live at \texttt{/testbed/AGENTS.md}. So anything the agent must know has to be \emph{in} that file or reachable from it. A brilliant script the agent is never pointed to changes nothing. A layer without \texttt{AGENTS.md} is rejected outright rather than run.

Your layer files sit in the repo untracked, so they never become part of the agent's submitted patch --- you cannot break its diff by adding files.
\end{itemize}

\textbf{How to work}

To see how the agent does on a task with your current environment, run:

{\fontsize{6}{7.5}\selectfont
\begin{verbatim}
python3 /work/box_run.py <instance_id> --layer /work/layer \
  --out /work/evolve_log/<name>
\end{verbatim}
}

It runs the fixed agent on that task inside your environment and writes the agent's full trajectory (\texttt{<out>/agent/trajectory.json}, containing every message and command) and its code change (\texttt{<out>/fix.diff}), and prints the run's \textbf{step count} (\texttt{n\_steps}) along with the paths to both. \textbf{Read the trajectory} --- it is your only signal. See where the agent wastes steps, and reshape the environment to remove that waste.

It also prints \texttt{exit\_status}, which says \emph{how} the run ended. Two values matter:

\begin{itemize}
\item \texttt{Submitted} --- the agent finished on its own terms.
\item anything else (e.g. hitting the 40-step limit) --- the agent ran out of road, and whatever it had done is thrown away.
\end{itemize}

The fix is read straight out of the repository when the run ends, so the agent does not have to hand anything over --- it finishes by running \texttt{echo COMPLETE\_TASK\_AND\_SUBMIT\_FINAL\_OUTPUT} on its own line. Whatever it has changed in the working tree counts, whether or not it got that far. So an empty \texttt{fix.diff} means the agent genuinely changed nothing, not that it fumbled a hand-off. Note only tracked source files count: scratch files it creates are ignored, as are yours.

\textbf{Also watch out for correctness --- do not let step reduction lead to any decrease in the resolve rate.} To judge whether a fix is right, don't rely on the step count, or on whether the fix's diff matches a previous run --- a matching diff only shows the agent did the same thing twice, not that it's correct. Check correctness yourself by reading the diff against the problem statement.

\textbf{You must do your optimization and tests on all the tasks you have, not just a handful of them.}

\textbf{Your budget for this step is \$30} of target-run spend. Keep a running total and stop before you exceed it.

\textbf{You are headless and non-interactive. There is NO one to re-invoke you and NO wakeups/notifications --- they do not fire here. The session lives ONLY while your turn is active: the moment you end your turn it ends permanently, and any run still going is lost to you.} So never end your turn to ``wait.'' Running across all the tasks can take a long time --- tens of minutes, sometimes hours --- and that is \textbf{normal, not stuck}. Stay in this one turn and wait it out: run in the \textbf{foreground} and let it finish, or if you background it, keep polling it yourself in the same turn until it is done. Only end your turn once you have finished iterating and written \texttt{REPORT.md}.

\textbf{Rules}

\begin{itemize}
\item \textbf{Do not lower how often the agent fixes bugs.} Fewer steps is worthless if it stops fixing them --- steps down, fixes intact.
\item \textbf{Any validation helper you build must fail when the tests it runs fail.} Never downgrade a failing test run to a weaker check such as an import check or a syntax check, never cache or suppress a previous pass, and never print a success message on a path where tests did not actually pass.
\item \textbf{Never cap how many times the agent may look.} Do not write fixed limits on searching, reading, or validating --- no ``exactly one search'', ``validate once then stop'', ``do not check twice''. A bug report names where the failure \emph{surfaced}, which is often not where the bug \emph{is}, so the agent must stay free to follow the trail to the real cause. Speed must come from pointing the agent at the right place, never from forbidding it a second look.
\item \textbf{You only ADD to the environment; the repo's existing source stays exactly as it is.} Every existing file, identifier, function signature, and error message must remain byte-for-byte unchanged --- do not rename, move, rewrite, overwrite, or delete existing code. (The bug and its hidden tests are pinned to that exact code; altering any of it corrupts the task.) Build your environment \emph{alongside} the code, never by editing it.
\item \textbf{Never put a corrected copy of this library, or anything derived from one, into the layer.} The agent's checkout has a single commit, no other branches and no network, so it cannot find one itself --- and you must not hand it one either, whether from a published release, a package index, or any other copy you can reach. This is deliberate: an answer read out of a fixed copy ``works'' on every task, including unseen ones, while measuring nothing, because a real bug has no corrected copy beside it. Build from the code as the agent finds it. A layer that ships or points at repaired source is discarded whatever it scores.
\item \textbf{Work within one task's checkout at a time.} Treat each task's source as the only source that exists while you are on it; never carry code from one task's checkout into another, or into the layer. Comparing \emph{bug reports} across tasks to find recurring themes is fine and encouraged. Carrying \emph{code} across them is not, and any layer containing source obtained that way is discarded whatever it scores.
\item \textbf{No cheating.} Don't swap in a different or stronger agent/model.
\item \textbf{Write about the codebase, not about your sample bugs.} Notes and cheat-sheets are welcome --- see ``How you are actually graded''. What is worthless is a line that encodes the answer to one task you were given: a check that looks for a particular corrected line, or a note stating what a particular function should return. Putting such a line under a heading like ``codebase invariants'', ``domain knowledge'' or ``navigation notes'' does not change what it is; the held-out bugs are in different functions and will not match it. Every line must earn its place on bugs you have never seen.
\item You never see the tests or the grade --- judge only from the agent's trajectories.
\end{itemize}

\textbf{When done}

Write \texttt{/work/REPORT.md}: what you reshaped, the waste you saw in the agent, and the effect on step count you observed.
\end{tcolorbox}

\section{Additional Results}
\label{appendix:additional_results}

Here we present results on the five benchmarks in addition to what is included in the main text.

\subsection{Detailed results on AutomationBench}

We show detailed performance of \textsc{\textsc{GEPA}}, Meta-Harness, \framework + Meta-Harness on tested agents in \cref{tab:detail_automationbench}.

\subsection{Validation set results}

We give the validation set results in \cref{tab:validation_results}. For each benchmark, we report the cost (or time for Bleve repository and number of steps for Tenacity repository, to be consistent with what we report in \cref{sec:experiments}) and pass rate of running each step's optimized environment with the corresponding target agent.

\begin{figure*}[t]
\centering
\includegraphics[width=0.8\textwidth]{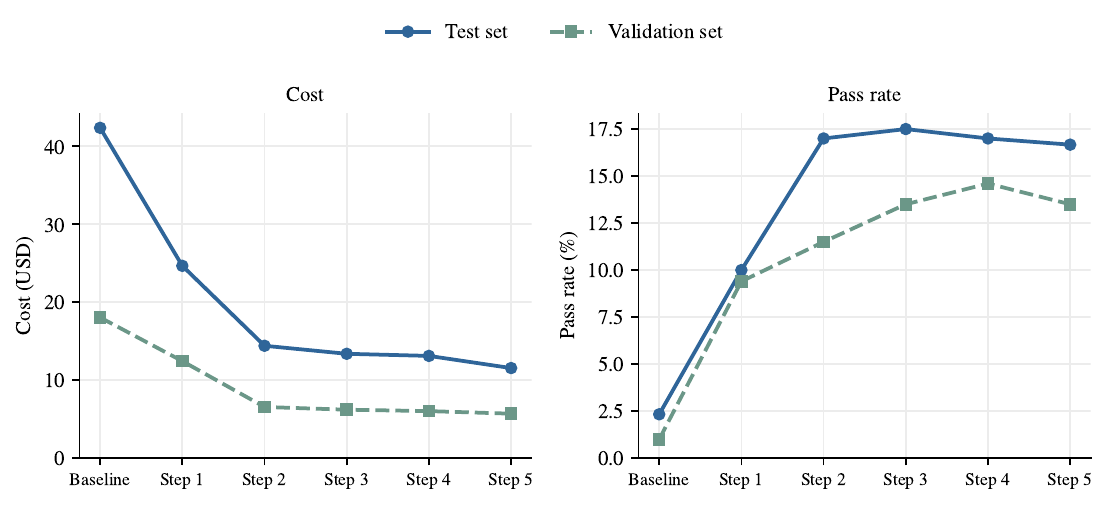}
\caption{
Cost (left) and pass rate (right) of \framework on AutomationBench after each step, on the test set and the validation set. The target agent is AutomationBench's official harness with Claude Haiku 4.5.
}
\label{fig:test_val_automationbench}
\end{figure*}

We also compare the results on test set and on validation set for AutomationBench, see \cref{fig:test_val_automationbench}.

\begin{table}[t]
\centering
\small
\caption{Pass rate and cost (USD) for each agent and optimization method on AutomationBench, averaged over 3 runs.}
\label{tab:detail_automationbench}
\begin{tabular}{llcc}
\toprule
Agent & Optimization & Pass (\%) & Cost (USD) \\
\midrule
\multirow{5}{*}{Haiku 4.5}
& Baseline                       & 2.33  & 42.36 \\
& \textsc{GEPA}                           & 1.83  & 39.45 \\
& Meta-Harness                   & 7.00  & 75.33 \\
& \framework                & 16.67 & 11.55 \\
& \framework + Meta-Harness & 18.17 & 12.47 \\
\midrule
\multirow{4}{*}{Haiku 5.5}
& Baseline                       & 20.7  & 4.83 \\
& Meta-Harness                   & 26.0  & 4.99 \\
& \framework                & 32.8  & 1.51 \\
& \framework + Meta-Harness & 32.8  & 1.49 \\
\midrule
\multirow{2}{*}{GPT-5.6-luna}
& Baseline                       & 19.17 & 17.42 \\
& \framework                & 24.67 & 1.98 \\
\midrule
\multirow{2}{*}{Opus 4.8}
& Baseline                       & 28.50 & 358.13 \\
& \framework                & 33.33 & 80.90 \\
\midrule
\multirow{4}{*}{Opus 5}
& Baseline                       & 33.83 & 236.18 \\
& Meta-Harness                   & 27.5  & 219 \\
& \framework                & 35.17 & 75.26 \\
& \framework + Meta-Harness & 35.8  & 76 \\
\bottomrule
\end{tabular}
\end{table}

\begin{table*}[t]
\caption{Validation set results for each step of \framework on each benchmark. The token efficiency metric differs by benchmark: cost for AutomationBench, Workspace-Bench and Data-Eng-Bench, time per bug for Bleve, and average agent steps for Tenacity.}
\label{tab:validation_results}
\vskip 0.15in
\begin{center}
\begin{small}
\begin{tabular}{lcccccccccc}
\toprule
 & \multicolumn{2}{c}{AutomationBench} & \multicolumn{2}{c}{Workspace-Bench} & \multicolumn{2}{c}{Data-Eng-Bench} & \multicolumn{2}{c}{Bleve} & \multicolumn{2}{c}{Tenacity} \\
\cmidrule(lr){2-3} \cmidrule(lr){4-5} \cmidrule(lr){6-7} \cmidrule(lr){8-9} \cmidrule(lr){10-11}
 & Acc. & Cost & Acc. & Cost & Acc. & Cost & Acc. & Time & Acc. & Steps \\
\midrule
Baseline & 1.0\%  & \$18.04 & 34.8\% & \$3.87 & 68.42\% & \$3.8952 & 96.7\%  & 116 s & 15.9\% & 24.20 \\
Step 1   & 9.4\%  & \$12.44 & 35.7\% & \$2.89 & 71.93\% & \$2.7132 & 90.0\%  & 46 s  & 18.2\% & 25.32 \\
Step 2   & 11.5\% & \$6.56  & 33.8\% & \$2.76 & 71.93\% & \$2.2104 & 96.7\%  & 54 s  & 18.2\% & 22.50 \\
Step 3   & 13.5\% & \$6.21  & 39.9\% & \$2.97 & 73.68\% & \$2.3078 & 96.7\%  & 52 s  & 13.6\% & 23.00 \\
Step 4   & 14.6\% & \$6.03  & 31.6\% & \$2.54 & 68.42\% & \$2.0018 & 95.0\%  & 53 s  & 18.2\% & 23.82 \\
Step 5   & 13.5\% & \$5.70  & 41.0\% & \$2.49 & 61.40\% & \$1.8106 & 96.7\%  & 52 s  & 11.4\% & 24.86 \\
Step 6   & 12.5\% & \$5.55  & 35.7\% & \$2.04 & ---     & ---      & 100.0\% & 57 s  & ---    & ---   \\
Step 7   & ---    & ---     & 40.5\% & \$2.12 & ---     & ---      & ---     & ---   & ---    & ---   \\
\bottomrule
\end{tabular}
\end{small}
\end{center}
\vskip -0.1in
\end{table*}

\subsection{How AutomationBench \framework converts a similar unseen environment}
\label{appendix:subsec:auto_convert}

For optimization over stationary environments, we do not require the environments to be exactly the same, slightly different yet highly similar environments can be converted using an automated converter written by the research agent during \framework. We give an example for AutomationBench, where the converter is a short program that runs at the start of every task. It calls the same API the agent uses to read that task's data, for example by listing all Help Scout conversations, and rewrites the results into one compact text block placed at the top of the agent's prompt.

In one environment, Help Scout holds 11 conversations. Without the converter, the agent must first search for the right endpoint, then call it, and then read records like this one:
{\scriptsize
  \begin{verbatim}
  {"id": "mig_hs1",
   "subject":
     "Invoice correction needed",
   "status": "active",
   "customer_id": "mig_cust1",
   "mailbox_id": "mb_main",
   "created_at": "2026-02-01T09:00:00Z",
   "tags": ["billing"],
   "threads": [{"id": "thr_1",
     "type": "customer",
     "body":
       "My last invoice was incorrect."}]}
  \end{verbatim}
  }
With the converter, the agent starts the task already holding every conversation, one line each, plus the exact call needed to act on them:

{\scriptsize
  \begin{verbatim}
  HELPSCOUT CONVERSATIONS (11 - full
    records; do NOT re-list/re-GET.
    WRITE an internal note: POST
    .../conversations/{ID}/threads/note)
  mig_hs1 | "Invoice correction needed"
    | status=active | tags=[billing]
    | mig_cust1 | customer: My last
    invoice was incorrect.
  mig_hs2 | "API rate limiting errors"
    | ...
  \end{verbatim}
  }
  
The agent therefore skips the search and listing steps and acts directly. On an unseen task with a different world, the same converter reads that world's data and builds its block in the same way.

\section{Example Synthetic Tasks}
\label{appendix:example_tasks}

We show some examples of the synthetic training tasks we used for Workspace-Bench and Data-Eng-Bench.

\subsection{Workspace-Bench}

For each of the five workspaces in Workspace-Bench, we showcase two example synthetic tasks in \cref{tab:wb-synth-examples}.

\begin{table*}[t] 
\centering
\small
\caption{Example synthetic training tasks for Workspace-Bench, two per workspace. Tasks are written without verifiers or reference answers.}
\label{tab:wb-synth-examples}
\begin{tabular}{p{0.11\textwidth} p{0.63\textwidth} p{0.2\textwidth}}
\toprule
Workspace & Task & Output file \\ 
\midrule
Product & Combine \texttt{Depreciation\_Breakdown\_2024\_1.csv} through \texttt{Depreciation\_Breakdown\_2024\_5.csv} into one spreadsheet, keeping a column that records which file each row came from. & \texttt{depreciation\_2024\_ consolidated.xlsx} \\
\cmidrule(l){2-3}
& Summarise the ten OKR files in the Product OKR folder, \texttt{\_01.txt} through \texttt{\_10.txt}, into one markdown file on the desktop. & \texttt{okr\_summary.md} \\
\midrule
Engineering & Index the ADR files in the architecture folder, from \texttt{ADR\_1.md} to \texttt{ADR\_11.md}: number, decision, and one line on what it constrains. Several are named only by number while others carry a topic in the filename, so rename them consistently and record
the old name against the new one. & \texttt{adr\_index.md} \\
\cmidrule(l){2-3}
& Extract the action items from \texttt{incident\_perf\_postmortem\_20250215.md} and say which have an owner. & \texttt{postmortem\_actions.md} \\
\midrule
Research & Read \texttt{Security\_Report\_1.txt} through \texttt{Security\_Report\_5.txt} in the 2024 folder and list the themes that appear in more than one. & \texttt{security\_themes.txt} \\
\cmidrule(l){2-3}
& Combine \texttt{2020-01\_Japanese\_Automobile\_Output\_by\_Model.xlsx} and \texttt{2020-01\_Japanese\_Automobile\_Sales\_by\_Model.xlsx} into one table by model. & \texttt{automaker\_2020.xlsx} \\
\midrule
Operations & Read the OKR target management and SQL basics training modules and say what prior knowledge each assumes. Then judge whether somebody joining the operations team could take them in either order or whether one depends on the other, and recommend a sequence for a new
starter with neither background. & \texttt{training\_module\_ notes.txt} \\
\cmidrule(l){2-3} 
& Summarise \texttt{user\_persona\_core\_user\_cohorts\_research\_2024.md} into a one-page brief for somebody about to write copy for these users. Keep the behavioural detail and drop the demographic padding, and mark any persona the research itself describes as provisional
rather than evidenced. & \texttt{persona\_brief.doc} \\
\midrule
Logistics & Sort the material\_requests folder into subfolders by status, renaming each file so its name carries the requester and the item, and leave a list of request, requester, item and status. & \texttt{material\_request\_ status.txt} \\
\cmidrule(l){2-3}
& Index the policies\_legal\_confirmed folder: policy name, what it governs, and its date. Sort the entries so the oldest sit at the top, since a legally confirmed policy that has not been revisited in years is the one most likely to have been overtaken, and flag any two
policies that appear to govern the same activity. & \texttt{policies\_index.md} \\
\bottomrule
\end{tabular} 
\end{table*}

\subsection{Data-Eng-Bench}

We show three examples of synthetic training tasks for Data-Eng-Bench in \cref{task:channel-mix,task:price-history,task:payroll}.

\begin{tcolorbox}[breakable, colback=gray!3, colframe=gray!50, boxrule=0.4pt,
                  arc=2pt, left=5pt, right=5pt, top=4pt, bottom=4pt,
                  fonttitle=\bfseries, parbox=false,
                  title={Example Training Task 1: dbt-order-channel-mix-synthetic}]
\refstepcounter{tasknum}\label{task:channel-mix}
\small
\sloppy
\textbf{Build Order Channel Mix Report}

\textbf{Overview}

The commercial team wants a monthly view of how order volume and revenue split across sales channels, so they can see which channels are growing and which are carrying discount-heavy demand.

\textbf{Files}

\begin{itemize}
\item DuckDB: \texttt{/\allowbreak app/\allowbreak dbt\_models\_duckdb/\allowbreak models/\allowbreak marts/\allowbreak sales/\allowbreak rpt\_order\_channel\_mix.sql}
\item Snowflake: \texttt{/\allowbreak app/\allowbreak dbt\_models\_snowflake/\allowbreak models/\allowbreak marts/\allowbreak sales/\allowbreak rpt\_order\_channel\_mix.sql}
\end{itemize}

\emph{[\ldots{} standard Database Backend / dbt Profile Setup boilerplate \ldots{}]}

\textbf{Requirements}

\textbf{Data Sources}

\begin{itemize}
\item \texttt{stg\_orders\_\_orders} - order header data; carries \texttt{channel\_id}, \texttt{ordered\_at}, \texttt{grand\_total}, \texttt{discount\_total}, \texttt{status}
\item \texttt{stg\_orders\_\_channels} - channel reference data (\texttt{channel\_id}, \texttt{channel\_code}, \texttt{channel\_name}, \texttt{channel\_type}, \texttt{is\_active})
\end{itemize}

\textbf{Grain}

One row per \texttt{(order\_month, channel\_id)}. \texttt{order\_month} is the calendar month of \texttt{ordered\_at}, formatted \texttt{YYYY-MM}.

\textbf{Scope Rules}

\begin{enumerate}
\item \textbf{Cancelled orders are excluded.} Filter out rows where \texttt{status = 'CANCELLED'}.
\item \textbf{Orders with a NULL \texttt{channel\_id} are excluded.} They cannot be attributed to a channel.
\item \textbf{Orders with a NULL \texttt{ordered\_at} are excluded.} They cannot be placed in a month.
\item \textbf{Inactive channels are still reported.} Join \texttt{stg\_orders\_\_channels} with a LEFT JOIN from the order side and keep channels where \texttt{is\_active} is false --- historic months legitimately contain orders on channels that have since been retired. When a \texttt{channel\_id} has no matching reference row, report \texttt{channel\_name} as \texttt{'UNKNOWN'} and \texttt{channel\_type} as \texttt{'UNKNOWN'}.
\end{enumerate}

\textbf{Metrics}

\begin{itemize}
\item \texttt{order\_count} - count of distinct \texttt{order\_id}
\item \texttt{customer\_count} - count of distinct non-NULL \texttt{customer\_id}
\item \texttt{gross\_revenue} - sum of \texttt{grand\_total}
\item \texttt{discount\_total} - sum of \texttt{discount\_total}
\item \texttt{avg\_order\_value} - \texttt{gross\_revenue / order\_count}, rounded to 2 decimals
\item \texttt{discount\_rate} - \texttt{discount\_total / gross\_revenue}, rounded to 4 decimals; return \texttt{0} when \texttt{gross\_revenue} is 0
\item \texttt{channel\_share\_of\_month} - this channel's \texttt{gross\_revenue} divided by the total \texttt{gross\_revenue} for the same \texttt{order\_month} across all channels, expressed as a percentage rounded to 2 decimals (e.g. \texttt{41.75})
\end{itemize}

\textbf{Expected Output}

The model produces these columns:

\begin{itemize}
\item \texttt{order\_month} - calendar month of the order, \texttt{YYYY-MM}
\item \texttt{channel\_id} - channel identifier
\item \texttt{channel\_name} - channel name, \texttt{'UNKNOWN'} when unmatched
\item \texttt{channel\_type} - channel type, \texttt{'UNKNOWN'} when unmatched
\item \texttt{order\_count} - distinct orders
\item \texttt{customer\_count} - distinct customers
\item \texttt{gross\_revenue} - sum of \texttt{grand\_total}
\item \texttt{discount\_total} - sum of \texttt{discount\_total}
\item \texttt{avg\_order\_value} - revenue per order
\item \texttt{discount\_rate} - discount as a share of gross revenue
\item \texttt{channel\_share\_of\_month} - percent of the month's revenue on this channel
\end{itemize}

Sort by \texttt{order\_month} ascending, then \texttt{channel\_id} ascending.

\textbf{Validation}

\begin{itemize}
\item Model compiles and runs successfully
\item Each \texttt{(order\_month, channel\_id)} pair appears exactly once
\item \texttt{channel\_share\_of\_month} sums to approximately 100 within each \texttt{order\_month}
\item No infinity or NaN values in any numeric column
\item Months with orders on retired channels still appear in the output
\end{itemize}

\textbf{Guidelines}

\begin{itemize}
\item The model should be materialized as a table
\item Do NOT modify upstream staging models
\item Keep the SQL portable across DuckDB and Snowflake
\end{itemize}
\end{tcolorbox}

\begin{tcolorbox}[breakable, colback=gray!3, colframe=gray!50, boxrule=0.4pt,
                  arc=2pt, left=5pt, right=5pt, top=4pt, bottom=4pt,
                  fonttitle=\bfseries, parbox=false,
                  title={Example Training Task 2: dbt-product-price-history-synthetic}]
\refstepcounter{tasknum}\label{task:price-history}
\small
\sloppy
\textbf{Build Product Price History Model}

\textbf{Overview}

Prices are held as effective-dated rows per variant, per price type, per currency, and optionally per customer tier and channel. There are several rows live at once for the same variant and nobody can say which one a customer actually saw, or how often prices move.

Build \texttt{rpt\_variant\_price\_movement}, a per-variant view of price movement and current position.

\textbf{Files}

\begin{itemize}
\item DuckDB: \texttt{/\allowbreak app/\allowbreak dbt\_models\_duckdb/\allowbreak models/\allowbreak marts/\allowbreak product/\allowbreak rpt\_variant\_price\_movement.sql}
\item Snowflake: \texttt{/\allowbreak app/\allowbreak dbt\_models\_snowflake/\allowbreak models/\allowbreak marts/\allowbreak product/\allowbreak rpt\_variant\_price\_movement.sql}
\end{itemize}

\emph{[\ldots{} standard Database Backend / dbt Profile Setup boilerplate \ldots{}]}

\textbf{Requirements}

\textbf{Data Sources}

\begin{itemize}
\item \texttt{stg\_product\_\_product\_prices} - \texttt{price\_id}, \texttt{variant\_id}, \texttt{price\_type}, \texttt{currency\_code}, \texttt{price\_amount}, \texttt{compare\_at\_price}, \texttt{cost\_price}, \texttt{min\_qty}, \texttt{max\_qty}, \texttt{customer\_tier\_id}, \texttt{channel\_id}, \texttt{effective\_from}, \texttt{effective\_to}, \texttt{is\_active}, \texttt{created\_by}
\item \texttt{stg\_product\_\_product\_variants} - \texttt{variant\_id}, \texttt{product\_id}, \texttt{sku}
\item \texttt{stg\_product\_\_products} - \texttt{product\_id}, \texttt{product\_name}, \texttt{primary\_category\_id}, \texttt{msrp}
\item \texttt{stg\_product\_\_product\_categories} - \texttt{category\_id}, \texttt{category\_name}
\end{itemize}

\textbf{The List Price}

Of the several price rows live for a variant, the \textbf{list price} is the one that applies to everybody: \texttt{price\_type} marking it as the standard list, \texttt{customer\_tier\_id} NULL, and \texttt{channel\_id} NULL. Tier-specific and channel-specific rows are negotiated or promotional and are counted separately rather than competing for the list position.

Inspect the \texttt{price\_type} vocabulary before deciding which value marks the list price. There is more than one type and the names are not self-evident.

\textbf{The Reference Date}

The latest \texttt{effective\_from} present anywhere in the price data. Do not use \texttt{CURRENT\_DATE}.

\textbf{Current Versus Historic}

A price row is \textbf{current} when it is active, its \texttt{effective\_from} is on or before the reference date, and its \texttt{effective\_to} is either NULL or on or after it.

Where two list-price rows are current for the same variant and currency, that is a configuration error. Take the one with the latest \texttt{effective\_from}, then the highest \texttt{price\_id}, and count the variant in \texttt{overlapping\_current\_prices}.

\textbf{Grain}

One row per \texttt{(variant\_id, currency\_code)} that has at least one list price row, current or historic.

\textbf{Price Movement}

Order the variant's list-price rows for that currency by \texttt{effective\_from}. A \textbf{change} is any consecutive pair where \texttt{price\_amount} differs. The size of a change is the new price minus the old, and its direction follows the sign.

Rows with a NULL \texttt{price\_amount} break the chain: exclude them from the movement calculation and count them in \texttt{undated\_or\_unpriced\_rows} along with rows carrying a NULL \texttt{effective\_from}.

\textbf{Output Columns}

{\scriptsize
\begin{tabular}{@{}l@{\hspace{4pt}}l@{\hspace{4pt}}p{0.42\linewidth}@{}}
\textbf{Column} & \textbf{Type} & \textbf{Description} \\
\hline
variant\_id & varchar & Variant identifier \\
sku & varchar & SKU \\
product\_id & varchar & Parent product \\
product\_name & varchar & Product name \\
category\_name & varchar & Primary category \\
currency\_code & varchar & Currency of the price series \\
list\_price\_rows & integer & List-price rows for this variant and currency \\
current\_list\_price & decimal & Resolved current price, NULL when none current \\
current\_effective\_from & date & When the current price took effect \\
first\_list\_price & decimal & Earliest priced row \\
first\_effective\_from & date & When it took effect \\
price\_changes & integer & Consecutive pairs that differ \\
increases & integer & Of those, upward \\
decreases & integer & Of those, downward \\
total\_change & decimal & Current minus first \\
total\_change\_pct & decimal & As a share of the first price \\
largest\_increase & decimal & Biggest single rise \\
largest\_decrease & decimal & Biggest single fall \\
avg\_days\_between\_changes & decimal & Mean gap between changes \\
days\_since\_last\_change & integer & Reference date minus the last change \\
tier\_price\_rows & integer & Tier-specific rows \\
channel\_price\_rows & integer & Channel-specific rows \\
current\_margin & decimal & Current price minus cost \\
current\_margin\_pct & decimal & Margin over current price \\
msrp\_gap & decimal & Product msrp minus current price \\
overlapping\_current\_prices & integer & 1 where competing current list rows existed \\
volatility & varchar & Classification \\
\end{tabular}}

Sort by \texttt{price\_changes} descending, then \texttt{sku} ascending.

\textbf{Success Criteria}

\begin{enumerate}
\item The model compiles and runs without errors
\item Each \texttt{(variant\_id, currency\_code)} pair appears exactly once
\item \texttt{increases} plus \texttt{decreases} equals \texttt{price\_changes} on every row
\item Exactly one current list price is resolved per row, or none where nothing is current
\item Variants whose only price rows are tier or channel specific do not appear
\item \texttt{volatility} is one of the four documented values on every row
\item No infinity or NaN values anywhere in the output
\end{enumerate}

\textbf{Guidelines}

\begin{itemize}
\item The model should be materialized as a table
\item Build the ordered price chain with a window function; a self-join on effective date will fan out wherever two rows share one
\item Do NOT modify upstream staging models
\end{itemize}
\end{tcolorbox}

\begin{tcolorbox}[breakable, colback=gray!3, colframe=gray!50, boxrule=0.4pt,
                  arc=2pt, left=5pt, right=5pt, top=4pt, bottom=4pt,
                  fonttitle=\bfseries, parbox=false,
                  title={Example Training Task 3: dbt-payroll-cost-by-department-synthetic}]
\refstepcounter{tasknum}\label{task:payroll}
\small
\sloppy
\textbf{Build Payroll Cost by Department Model}

\textbf{Overview}

Payroll is run in periods and posted as a run header with employee-level detail beneath it. Time is recorded separately, in entries per employee per day. The two are never compared, so nobody can say whether the hours being paid match the hours being worked, or what a department actually costs.

Build \texttt{rpt\_payroll\_cost\_by\_department}, which brings the three together.

\textbf{Files}

\begin{itemize}
\item DuckDB: \texttt{/\allowbreak app/\allowbreak dbt\_models\_duckdb/\allowbreak models/\allowbreak marts/\allowbreak hr/\allowbreak rpt\_payroll\_cost\_by\_department.sql}
\item Snowflake: \texttt{/\allowbreak app/\allowbreak dbt\_models\_snowflake/\allowbreak models/\allowbreak marts/\allowbreak hr/\allowbreak rpt\_payroll\_cost\_by\_department.sql}
\end{itemize}

\emph{[\ldots{} standard Database Backend / dbt Profile Setup boilerplate \ldots{}]}

\textbf{Requirements}

\textbf{Data Sources}

\begin{itemize}
\item \texttt{stg\_hr\_\_payroll\_runs} - \texttt{payroll\_run\_id}, \texttt{payroll\_period}, \texttt{period\_start}, \texttt{period\_end}, \texttt{pay\_date}, \texttt{status}, \texttt{total\_gross}, \texttt{total\_net}, \texttt{employee\_count}
\item \texttt{stg\_hr\_\_payroll\_details} - \texttt{detail\_id}, \texttt{payroll\_run\_id}, \texttt{employee\_id}, \texttt{gross\_pay}, \texttt{tax\_deductions}, \texttt{other\_deductions}, \texttt{net\_pay}, \texttt{hours\_worked}, \texttt{overtime\_hours}
\item \texttt{stg\_hr\_\_time\_entries} - \texttt{entry\_id}, \texttt{employee\_id}, \texttt{entry\_date}, \texttt{clock\_in}, \texttt{clock\_out}, \texttt{break\_minutes}, \texttt{hours\_worked}, \texttt{entry\_type}, \texttt{status}
\item \texttt{stg\_hr\_\_employees} - \texttt{employee\_id}, \texttt{department\_id}, \texttt{hire\_date}, \texttt{termination\_date}, \texttt{employment\_type}, \texttt{status}
\item \texttt{stg\_hr\_\_departments} - \texttt{department\_id}, \texttt{department\_code}, \texttt{department\_name}
\end{itemize}

\textbf{Grain}

One row per \texttt{(department\_code, payroll\_period)}.

\textbf{Only Posted Runs}

Payroll runs that were drafted and never posted must be excluded from every figure. Inspect the run \texttt{status} vocabulary before deciding which mean posted; there is more than one non-posted value.

Detail rows whose \texttt{payroll\_run\_id} matches no run are orphans and are excluded, counted in \texttt{orphan\_detail\_rows}.

\textbf{Matching Time To Pay}

For each detail row, sum the employee's time entries whose \texttt{entry\_date} falls within the run's \texttt{period\_start} to \texttt{period\_end} inclusive. Only entries in an approved state count; inspect the entry \texttt{status} values.

The comparison is between \texttt{hours\_worked} on the payroll detail and the summed \texttt{hours\_worked} from the time entries. A difference of more than 0.5 hours is a \textbf{discrepancy}. Report the count and the net direction, because paying more hours than were recorded and recording more than were paid are different problems with different owners.

Employees with payroll detail and no time entries at all in the period are counted separately from those with a discrepancy. A salaried employee who does not clock in is normal; a waged one who does not is not, and \texttt{employment\_type} is what distinguishes them.

\textbf{Output Columns}

\begin{itemize}
\item \texttt{department\_code}, \texttt{department\_name}, \texttt{payroll\_period}
\item \texttt{period\_start}, \texttt{period\_end}, \texttt{pay\_date}
\item \texttt{employees\_paid} - distinct employees with a detail row
\item \texttt{gross\_pay}, \texttt{tax\_deductions}, \texttt{other\_deductions}, \texttt{net\_pay} - summed
\item \texttt{deduction\_rate} - deductions over gross, rounded to 4 decimals
\item \texttt{paid\_hours}, \texttt{paid\_overtime\_hours} - summed from payroll detail
\item \texttt{recorded\_hours} - summed from approved time entries in the period
\item \texttt{hours\_gap} - \texttt{paid\_hours} minus \texttt{recorded\_hours}
\item \texttt{employees\_with\_discrepancy}
\item \texttt{employees\_overpaid\_hours}, \texttt{employees\_underpaid\_hours} - the split by direction
\item \texttt{employees\_without\_time\_entries}
\item \texttt{salaried\_without\_time\_entries} - of those, on a salaried employment type
\item \texttt{avg\_gross\_per\_employee}, \texttt{avg\_hourly\_rate} - the second \texttt{gross\_pay} over \texttt{paid\_hours}, both rounded to 2 decimals
\item \texttt{overtime\_share} - overtime hours over paid hours, rounded to 4 decimals
\item \texttt{header\_gross\_drift} - the run header's \texttt{total\_gross} minus the summed detail \texttt{gross\_pay}, apportioned to this department by its share of the run's detail rows and rounded to 2 decimals
\item \texttt{orphan\_detail\_rows} - repeated on every row
\end{itemize}

Employees whose department does not resolve group under \texttt{'UNKNOWN'}.

Sort by \texttt{payroll\_period} ascending, then \texttt{gross\_pay} descending.

\textbf{Why Three Sources And Not Two}

Payroll detail alone gives cost. Time entries alone give effort. Neither on its own answers the question a department head actually asks, which is whether the cost is buying the effort.

The run header is the third source and it exists for a different reason: it is what the finance system was told. Where the header disagrees with the detail beneath it, the general ledger and this model will report different numbers for the same period, and somebody will spend a day finding out why. \texttt{header\_gross\_drift} is there so that day is not spent.

The apportionment of that drift by share of detail rows is deliberately crude. A drift on the header cannot be attributed to a department, because the header does not know about departments. Spreading it evenly by row count is a convention, not a measurement, and the column should be read as ``roughly this much of the unexplained difference sits with this department'' rather than as a figure to book.

\textbf{Notes On The Time Data}

Time entries carry an \texttt{entry\_type} alongside their status, and the two are independent. An entry can be approved and still be of a type that should not count toward paid hours --- training, unpaid leave, a correction entry that zeroes an earlier one.

Inspect the \texttt{entry\_type} values before deciding which contribute to \texttt{recorded\_hours}. Including every approved entry regardless of type will inflate recorded hours and turn genuine overpayments into apparent underpayments, which is worse than reporting nothing.

Where an entry type is ambiguous, include it and say so in a comment. An inflated \texttt{recorded\_hours} produces conservative discrepancy counts, and under-reporting a discrepancy is the safer error here: it sends fewer people to investigate things that turn out to be fine.

Break minutes are recorded separately and are already excluded from \texttt{hours\_worked} on the entry. Do not subtract them again.

\textbf{Worked Example}

A payroll period 2023-07-B running 2023-07-16 to 2023-07-31, posted, with three employees in the Fulfilment department.

Employee e-10, waged, has payroll detail showing 72.0 hours and 6.0 overtime, gross 1,560.00. Their approved time entries in the window sum to 71.5 hours. The gap is 0.5, which is not more than 0.5, so it is not a discrepancy.

Employee e-11, waged, shows 80.0 paid hours. Their approved entries sum to 68.0. The gap is 12.0 and counts as a discrepancy in the overpaid direction --- twelve hours were paid that nobody recorded.

Employee e-12, salaried, shows 75.0 paid hours and has no time entries at all. They count in \texttt{employees\_without\_time\_entries} and in \texttt{salaried\_without\_time\_entries}, and not in \texttt{employees\_with\_discrepancy}. A salaried employee not clocking in is expected; the model separates them precisely so the waged case stands out.

So for that department and period: \texttt{employees\_paid} 3, \texttt{paid\_hours} 227.0, \texttt{recorded\_hours} 139.5, \texttt{hours\_gap} 87.5, \texttt{employees\_with\_discrepancy} 1, \texttt{employees\_overpaid\_hours} 1, \texttt{employees\_underpaid\_hours} 0, \texttt{employees\_without\_time\_entries} 1, \texttt{salaried\_without\_time\_entries} 1.

Note that \texttt{hours\_gap} of 87.5 is dominated by the salaried employee, whose 75 hours have no time entries by design. The gap column is therefore not a discrepancy measure on its own, and reading it as one would flag every department carrying salaried staff. \texttt{employees\_with\_discrepancy} is the measure that means something.

\textbf{Edge Cases}

An employee appearing in two departments across two periods produces a row in each. The employee record carries only the current department, so both periods are attributed to it --- that limitation is the same one the headcount model has and is worth the same comment.

An employee with detail rows in two runs covering overlapping periods will have their time entries counted against both. Overlapping payroll periods are a data error rather than a normal occurrence; count the affected employees in \texttt{employees\_in\_overlapping\_periods}, reported on every row, and leave the double count in place rather than trying to apportion it.

A time entry whose \texttt{clock\_out} is before its \texttt{clock\_in} crosses midnight or is corrupt. Trust \texttt{hours\_worked} on the entry rather than recomputing it from the two timestamps --- the entry's own figure is what the approval process signed off.

A detail row with a NULL \texttt{hours\_worked} belongs to a salaried employee on a period-based arrangement. It contributes to pay and to no hours figure.

A run whose header \texttt{employee\_count} disagrees with the count of detail rows is a header drift like any other. It is not reported here; the money columns are what matter and \texttt{header\_gross\_drift} covers them.

\textbf{Success Criteria}

\begin{enumerate}
\item The model compiles and runs without errors
\item Each \texttt{(department\_code, payroll\_period)} pair appears exactly once
\item \texttt{employees\_overpaid\_hours} plus \texttt{employees\_underpaid\_hours} equals \texttt{employees\_with\_discrepancy}
\item Unposted payroll runs contribute to no figure
\item \texttt{net\_pay} plus \texttt{tax\_deductions} plus \texttt{other\_deductions} equals \texttt{gross\_pay} on every row, within a cent
\item \texttt{employees\_without\_time\_entries} never exceeds \texttt{employees\_paid}
\item No infinity or NaN values anywhere in the output
\end{enumerate}

\textbf{Guidelines}

\begin{itemize}
\item The model should be materialized as a table
\item Aggregate the time entries to one row per employee and run before joining them to the payroll detail. Joining entries directly will multiply the pay by the number of days worked, which is the largest error this model can produce and the least obvious.
\item Do not use \texttt{CURRENT\_DATE}; the periods carry their own dates
\item Do NOT modify upstream staging models
\end{itemize}
\end{tcolorbox}

\end{document}